\documentclass[10pt,twocolumn,letterpaper]{article}

\usepackage{cvpr}              % To produce the CAMERA-READY version
\usepackage{booktabs}
\usepackage{multicol}
\usepackage{multirow}
\definecolor{cvprblue}{rgb}{0.21,0.49,0.74}
\usepackage[pagebackref,breaklinks,colorlinks,allcolors=cvprblue]{hyperref}
\usepackage[capitalize]{cleveref}
\def\paperID{11892} % *** Enter the Paper ID here
\def\confName{CVPR}
\def\confYear{2025}

\title{Fair Distillation: Teaching Fairness from Biased Teachers in Medical Imaging}

\author{
\parbox{0.7\linewidth}{\centering
Milad Masroor \textsuperscript{\rm 1} $\thanks{First two authors contributed equally to this work.}$ $\quad$
Tahir Hassan \textsuperscript{\rm 1} \footnotemark[1] $\quad$
Yu Tian \textsuperscript{\rm 2} $\quad$
Kevin Wells \textsuperscript{\rm 1} $\quad$
David Rosewarne \textsuperscript{\rm 1,3} $\quad$
Thanh-Toan Do \textsuperscript{\rm 4} $\quad$
Gustavo Carneiro\textsuperscript{\rm 1} $\newline$ 
\textsuperscript{\rm 1} CVSSP, PAI, University of Surrey, UK \\
\textsuperscript{\rm 2} University of Pennsylvania, USA \\
\textsuperscript{\rm 3} Royal Wolverhampton Hospitals NHS Trust, UK \\
\textsuperscript{\rm 4} Department of Data Science and AI, Monash University, Australia}
}

\begin{document}
\maketitle
\begin{abstract}
%While deep learning has produced outstanding results for multiple classification tasks, concerns over fairness remain, particularly as models may exhibit biases that impact sensitive demographic groups based on attributes like race, gender, or age. Current bias-mitigation techniques, such as Subgroup Re-balancing, Adversarial Training, and Domain Generalization, aim to balance accuracy across various demographic groups but struggle to simultaneously improve overall accuracy, group-specific accuracy, and fairness due to entanglement of these conflicting objectives. We propose Biased Teachers are Great Role Models for Unbiased Students (FairDi), the first fairness method designed to disentangle these objectives. Our approach initially trains biased ``teacher'' models, each optimized for a specific sensitive demographic group, which are sub-sequentially used to guide the training of a unified ``student'' model, which distills the knowledge from the teacher models, while maximizing overall accuracy and minimizing inter-group disparities. Evaluations on medical imaging datasets show that FairDi significantly improves overall and group-specific accuracy and fairness over existing methods. Adaptable to various medical tasks, such as classification and segmentation, our approach offers a competitive solution for equitable model performance, particularly in healthcare.
Deep learning has achieved remarkable success in image classification and segmentation tasks. However, fairness concerns persist, as models often exhibit biases that disproportionately affect  demographic groups defined by sensitive attributes such as race, gender, or age. Existing bias-mitigation techniques, including Subgroup Re-balancing, Adversarial Training, and Domain Generalization, aim to balance accuracy across demographic groups, but often fail to simultaneously improve overall accuracy, group-specific accuracy, and fairness due to conflicts among these interdependent objectives. We propose the \textbf{Fair} \textbf{Di}stillation (FairDi) method, a novel fairness approach that decomposes these objectives by leveraging biased ``teacher'' models, each optimized for a specific demographic group. These teacher models then guide the training of a unified ``student'' model, which distills their knowledge to maximize overall and group-specific accuracies, while minimizing inter-group disparities. Experiments on medical imaging datasets show that FairDi achieves significant gains in both overall and group-specific accuracy, along with improved fairness, compared to existing methods. FairDi is adaptable to various medical tasks, such as classification and segmentation, and provides an effective solution for equitable model performance.
%, especially in healthcare settings.
\end{abstract}
\section{Introduction}
\label{sec:intro}

%\textcolor{red}{We need to change a bit the narrative.  Before, we were saying that there was a tradeoff between overall AUC and gap, but that is not true.  As the gap reduces between cohorts, the overall AUC tends to increase.  So it is becoming clearer that it is important to improve as much as possible the performance per cohort (by biasing cohort-specific teachers) and somehow distill the information from the biased teachers.}

In recent years, deep learning models have made remarkable progress, offering automated solutions for tasks such as classification and segmentation. However, as these models transition into practice, 
%such as with medical image analysis methods getting deployed into clinical settings, 
concerns regarding fairness have emerged~\cite{obermeyer2019dissecting, mehrabi2021survey, kadambi2021achieving, chen2021ethical, chen2023algorithmic}. Machine learning models, particularly those used in healthcare, have been shown to exhibit biases toward demographic groups defined by sensitive attributes such as race, gender, and age~\cite{zong2022medfair, dutt2023fairtune, han2024ffb}. These biases can lead to unfair treatment of specific groups, exacerbating existing disparities. 
%in healthcare. 
Addressing these biases is not only a technical challenge but also a moral and ethical imperative to ensure equitable healthcare for all individuals.

% \begin{figure}[h!]
%   \centering
%   \includegraphics[width=\linewidth]{Figures/Motivation_figure.png}
%   \caption{This figure illustrates the negative correlation between
% performance (Overall AUC) and fairness (AUC Gap) for various
% models (ERM~\cite{vapnik1999overview}, GroupDRO~\cite{sagawa2019distributionally}, SWAD~\cite{cha2021swad}, FIS~\cite{luo2024fairvisionequitabledeeplearning} and ours) evaluated on the HAM10000 dataset by age (left) and gender (right) cohorts.
% Our approach shows superior results with both high overall AUC
% and low AUC Gap, suggesting equitable performance across age/gender subgroups. \textcolor{red}{Waiting for graph showing the group-specific AUC with 3 different ways to do model selection.}} 
%    \label{fig:motivation}
% \end{figure}

\begin{figure}[t!]
  \centering
  \includegraphics[width=\linewidth]{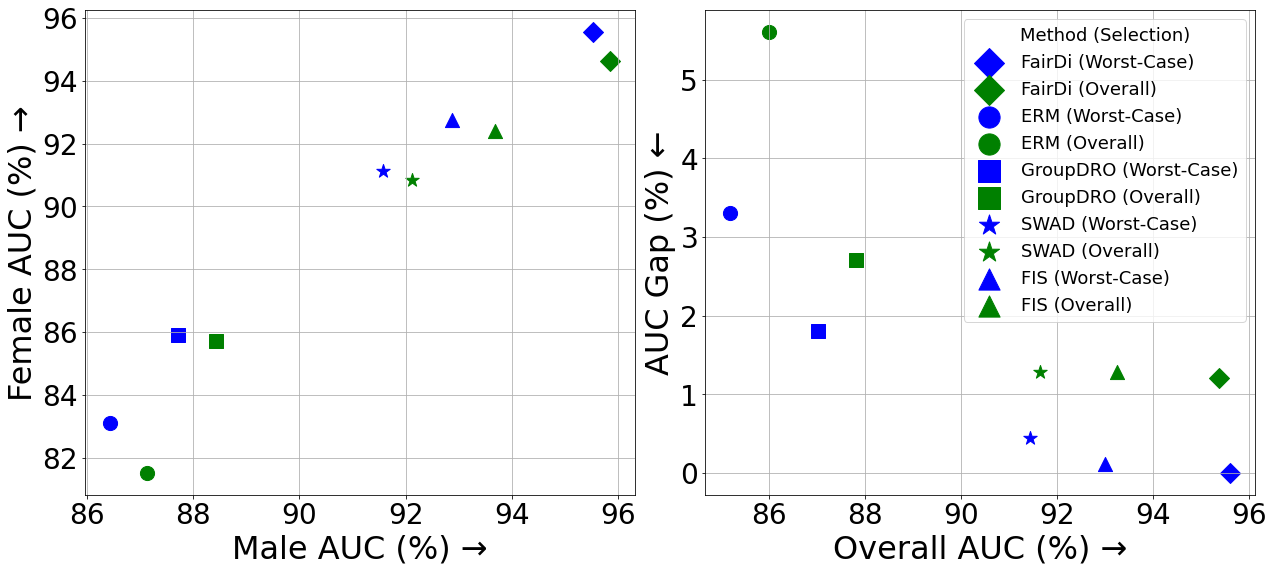}
  \caption{Performance comparison of models (ERM~\cite{vapnik1999overview}, GroupDRO~\cite{sagawa2019distributionally}, SWAD~\cite{cha2021swad}, FIS~\cite{luo2024fairvisionequitabledeeplearning}, and our FairDi) on the HAM10000 dataset~\cite{tschandl2018ham10000} for benign/malignant classification by gender. The left panel shows group-specific AUCs (Male and Female), and the right panel plots fairness (AUC Gap) vs. Overall AUC. Each model's Pareto front includes two points: one maximizing worst-group AUC, the other maximizing overall AUC. Our FairDi achieves the best balance with high overall AUC, low AUC Gap, and robust group-specific AUCs.} 
   \label{fig:motivation}
\end{figure}

%Current approaches to mitigating bias, including Subgroup Re-balancing~\cite{nitesh2002smote, idrissi2022simple}, Adversarial Training~\cite{madras2018learning, zhao2019conditional}, and Domain Generalization~\cite{cha2021swad, sagawa2019distributionally}, aim to reduce disparities between the performance of sensitive groups by modifying the training process or adjusting data distributions. These methods typically seek to maximize empirical accuracy for each sensitive group, while maximizing overall accuracy and minimizing accuracy gaps between groups. Consequently, these approaches can be formulated as multi-objective optimization problems (MOOPs)~\cite{martinez2020minimax}, where solutions aim to achieve the best possible trade-offs among objectives. Such solutions form a set of solutions that are non-dominated to each other but are superior to the rest of solutions in the search space. This means that for any solution in this set,  improving one objective (e.g., maximizing overall accuracy) often negatively impacts others (e.g., widening the accuracy gap between sensitive groups), resulting in a solution that lies on the Pareto front~\cite{zong2022medfair,MasColell1995microeconomic,martinez2020minimax}. For fairness methods, each method will have its own Pareto front that represents the optimal trade-offs among objectives.

Current bias mitigation approaches~\cite{nitesh2002smote, idrissi2022simple,madras2018learning, zhao2019conditional,tartaglione2021end,sarhan2020fairness,cha2021swad, sagawa2019distributionally,foret2020sharpness} 
%, such as Subgroup Re-balancing~\cite{nitesh2002smote, idrissi2022simple}, Adversarial Training~\cite{madras2018learning, zhao2019conditional}, Disentanglement~\cite{tartaglione2021end,sarhan2020fairness}, and Domain Generalization~\cite{cha2021swad, sagawa2019distributionally,foret2020sharpness}, 
aim to reduce performance disparities across sensitive groups by modifying the training process or adjusting data distributions. These methods typically seek to maximize empirical accuracy for each sensitive group, while also improving overall accuracy and minimizing accuracy gaps between groups. Consequently, they can be formulated as multi-objective optimization problems~\cite{martinez2020minimax}, with solutions designed to achieve optimal trade-offs among conflicting objectives.
These trade-offs form a set of non-dominated solutions, which are superior to all other solutions in the search space but cannot be further improved in one objective without compromising another. In this context, each fairness method generates a unique Pareto front, representing its best possible balance among objectives such as overall accuracy and inter-group fairness~\cite{zong2022medfair, MasColell1995microeconomic, martinez2020minimax}.  

\cref{fig:motivation} compares fairness methods ERM~\cite{vapnik1999overview}, GroupDRO~\cite{sagawa2019distributionally}, SWAD~\cite{cha2021swad}, FIS~\cite{luo2024fairvisionequitabledeeplearning}, and our FairDi on the HAM10000 dataset~\cite{tschandl2018ham10000} for benign/malignant classification by gender. The left panel illustrates group-specific AUCs (Male and Female), while the right panel plots AUC Gap versus Overall AUC. Each model’s Pareto front includes points for maximizing either worst-group or overall AUC, demonstrating trade-offs between fairness and accuracy. Unlike prior methods, which typically sacrifice overall or group-specific AUC for fairness, FairDi applies a novel decomposed optimization of the Pareto front, allowing it to achieve a superior balance, with high overall AUC, low AUC Gap, and robust group-specific AUCs, outperforming other methods on these metrics.

In this paper, we introduce the first fairness method, titled \textbf{Fair} \textbf{Di}stillation (FairDi), which is specifically designed to decompose the objectives of maximizing accuracy for each sensitive group from maximizing overall accuracy and minimizing inter-group accuracy gaps.
This objective decomposition is achieved in two stages: first, we train highly biased models, each optimized to maximize accuracy within a specific sensitive group. These models then serve as ``teachers'' for a unified ``student'' model, which distills knowledge from the biased ``teachers'', while simultaneously training to maximize overall accuracy and reduce accuracy gaps across groups, effectively balancing fairness and performance.
The main contributions of this paper are:
\begin{itemize}
    \item Introduction of the first fairness method, called FairDi, that decomposes accuracy maximization for each sensitive group from  overall accuracy maximization and inter-group accuracy gap minimization, with a teacher-student model, where biased teachers are trained for specific groups, while a single student distills their knowledge, maximizing overall accuracy and reducing accuracy gaps across groups;
    \item Proposal of a new teacher-student training algorithm that is highly adaptable to various datasets and both classification and segmentation tasks.
\end{itemize}
Comprehensive evaluations across diverse medical imaging datasets show that FairDi significantly outperforms state-of-the-art fairness methods, achieving higher overall and group-specific accuracies while minimizing inter-group disparities. Its adaptability to various tasks, including classification and segmentation, underscores its potential to advance equitable AI in critical applications.

\section{Related Work}
\label{sec:Related}

\textbf{Fairness in Artificial Intelligence (AI)}, particularly in healthcare, has gained increasing attention as machine learning models have been shown to exhibit biases related to sensitive attributes such as race, gender, and age~\cite{quadrianto2019discovering,ramaswamy2021fair,zhang2020towards,park2022fair,roh2020fr,sarhan2020fairness,zafar2017fairness,zhang2018mitigating,wang2022fairness,kim2019multiaccuracy,lohia2019bias,tian2024fairdomain,tian2024fairseg,luo2024fairclip,luo2023harvard,luo2023glau_fair,luo2023eye,zong2022medfair,shi2023equitable,shi2024equitable}. Studies by \citet{obermeyer2019dissecting} and \citet{larrazabal2020gender} highlight how biases in healthcare models can lead to disparities in outcomes, particularly in underrepresented populations, making fairness an essential concern in the development of AI systems for healthcare.

Bias mitigation techniques to address fairness concerns can be categorized into pre-processing, in-processing, and post-processing strategies. Pre-processing methods, like SMOTE~\cite{nitesh2002smote}, aim to re-balance data before training, while post-processing approaches~\cite{pleiss2017fairness} adjust model predictions based on sensitive attributes after training. Focusing on in-processing methods, one prominent method is Adversarial Training~\cite{zhang2018mitigating, kim2019learning}, which minimizes bias by optimizing the model to perform well on the classification task while reducing its ability to predict sensitive attributes. 
%This approach has been applied in several studies to enhance fairness by neutralizing the influence of sensitive attributes in learned representations. 
Despite their success, these methods often encounter a trade-off between fairness and model performance~\cite{madras2018learning, zhao2019conditional}.
Another notable in-processing method is Group Distributionally Robust Optimization (GroupDRO)~\cite{sagawa2019distributionally} that focuses on minimizing worst-case performance across demographic groups. Although GroupDRO ensures fairness across underperforming groups, it may reduce overall model performance~\cite{zong2022medfair}.

Domain generalization techniques, e.g., Stochastic Weight Averaging Densely (SWAD)~\cite{cha2021swad}, represent in-processing methods that have shown promising results, with a training process that finds flat minima in the loss landscape. However, despite its generally robust performance, SWAD may inadvertently widen disparities between subgroups, causing it to not significantly outperform the baseline Empirical Risk Minimization (ERM) method, raising questions about its suitability as a standalone solution for fairness~\cite{zong2022medfair}. Another relevant in-processing method is based on representation disentanglement approaches that aim to separate sensitive attributes from task-relevant information in the learned representations. \citet{tartaglione2021end} and~\citet{sarhan2020fairness} proposed disentanglement techniques that isolate sensitive attributes from the representation space, ensuring that the downstream task is unaffected by these attributes. However, while disentanglement can be effective in isolating individual sources of bias, it often underperforms in complex tasks where multiple biases coexist. This narrow focus limits generalization and efficacy, where confounding factors are deeply intertwined and challenging to separate. 

%Despite the range of bias mitigation techniques available, most of these approaches do not significantly outperform ERM in practice. Many of these methods, including domain generalization and representation disentanglement, were initially developed with objectives other than fairness and are therefore not explicitly designed to tackle group fairness issues. As a result, they do not satisfactorily address the unique fairness challenges posed by sensitive demographic groups, especially in the context of medical imaging~\cite{dutt2023fairtune}.

Recent fairness methods introduce bias mitigation approaches to address group fairness in conjunction with overall accuracy. For instance, FairSeg~\cite{tian2024fairseg} proposes Fair Error-Bound Scaling, which adjusts loss functions to focus on the hardest cases within each demographic group, ensuring equitable performance across different sensitive attributes. FairCLIP~\cite{luo2024fairclip} reduces bias in vision-language models using an optimal transport-based method to align demographic distributions, while FairVision~\cite{luo2024fairvisionequitabledeeplearning} introduces Fair Identity Scaling (FIS) to improve fairness in both 2D and 3D medical imaging. Importantly, all these methods utilize a single model, which often suffers from the conflicting objectives of group fairness and overall performance. In contrast, our approach employs cohort-specific teacher models that optimize performance independently for each cohort before distilling their knowledge into a unified student model, thus addressing fairness across groups without compromising overall performance. 

\textbf{Knowledge distillation} has become a powerful technique in machine learning, with significant applications across various domains~\cite{gou2021knowledge}. Firstly introduced by~\citet{hinton2015distilling,buciluǎ2006model}, knowledge distillation is widely used for model compression~\cite{sanh2019distilbert}, semi-supervised learning~\cite{tarvainen2017mean}, and multi-modal learning~\cite{lu2019vilbert}.
We propose an extension of knowledge distillation to the realm of fairness in AI. By leveraging knowledge distillation to transfer cohort-specific knowledge from multiple teacher models to a unified student model, we present the first approach to utilize knowledge distillation explicitly for achieving fairness across demographic groups and optimizing model overall performance.

\section{Method}
\label{sec:meth}

%\subsection{Knowledge Distillation for Fairness: Overview}

\begin{figure}[t]
  \centering
  \includegraphics[width=\linewidth]{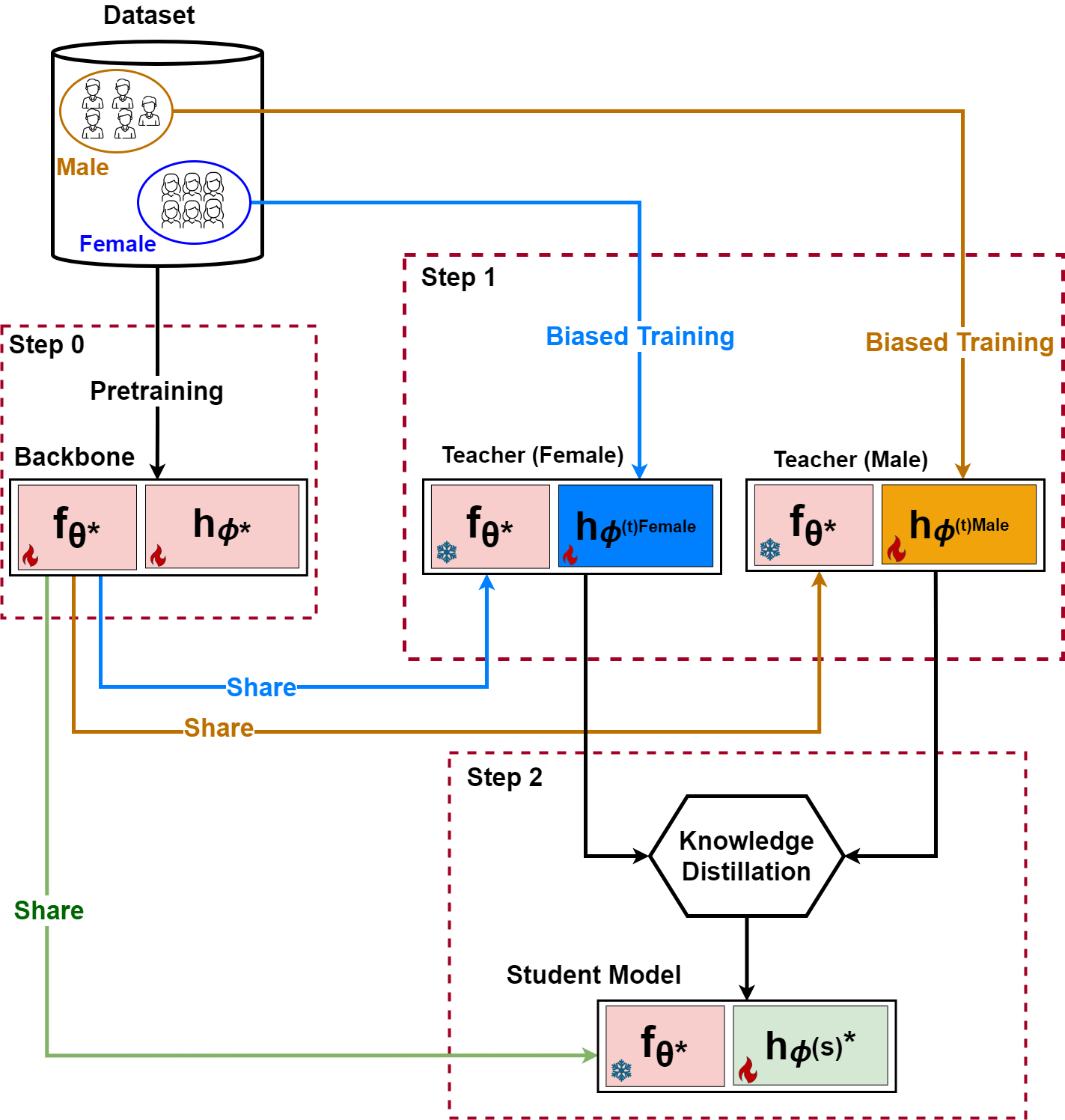}
   \caption{Diagram of the FairDi training process. After pre-training (step 0), it optimizes the teachers' performances for their respective sensitive groups (step 1), followed by a  knowledge distillation process that trains a single student to optimize overall and group-specific accuracies, while minimizing performance gap between groups (step 2). The \textcolor{red}{flame} symbol indicates model components to be trained during each corresponding step, while the \textcolor{blue}{snowflake} symbol represents components that remain frozen.}
   \label{fig:algorithm}
\end{figure}

Our proposed method addresses fairness in medical image analysis by implementing a student-teacher knowledge-distillation framework designed to balance cohort-specific accuracy with overall accuracy and fairness among demographic groups, as shown in Fig.~\ref{fig:algorithm}. 
The approach begins by pre-training a backbone model on the full dataset using a loss function~\cite{luo2024fairvisionequitabledeeplearning} that jointly optimizes the conflicting objectives of overall accuracy and fairness (Step 0). While this initial pre-training struggles to achieve an ideal balance between these objectives, it yields a robust backbone model that serves as a strong foundation for the subsequent cohort-specific teacher models and the unified student model. 
We then create one teacher model per demographic cohort (e.g., male or female) using the pre-trained backbone, which is greedily fine-tuned to optimize  cohort-specific accuracy (Step 1).
In the final stage (Step 2), our framework distills knowledge from the biased teacher models into a single student model, aiming to jointly optimize group-specific and overall accuracies together with fairness. This approach seeks to achieve the highest possible cohort-specific accuracies while also maximizing overall accuracy and fairness for the resulting model.
%which integrates and balances the cohort-specific insights. This distillation process allows the student model to minimize disparities across demographics, creating a fairer representation that mitigates common trade-offs between accuracy and fairness. Our approach demonstrates that high overall accuracy and demographic equity are not conflicting goals but can be achieved concurrently.

\subsection{Backbone Model Training (Step 0)}
\label{sec:training_step_0}

We assume to have a labeled dataset $\mathcal{D} = \{ (x_i,y_i,a_i) \}_{i=1}^{N}$, where $x \in \mathcal{X} \subset \mathbb{R}^{W \times H \times R}$ is an RGB image, $y \in \mathcal{Y}$ denotes the training label, with $\mathcal{Y} \subset \{0,1\}^{Y}$ for classification and $\mathcal{Y} \subset \{0,1\}^{W \times H \times Y}$ for segmentation,
and
$a \in \mathcal{A} \subset \{0,\cdots,A\}$ represents 
the value of the identity attribute (e.g., for the gender attribute, the values can be male or female) associated with the $i^{th}$ sample of the dataset $\mathcal{D}$. 
Since we use stochastic gradient descent (or a variant) for optimization, $\mathcal{D}$ in the formulations below denotes a mini-batch rather than the entire training set.
The backbone model transforms a sample from the image space to the feature space $\mathcal{F}$ with $f_{\theta}:\mathcal{X} \to \mathcal{F}$ parameterized by $\theta \in \Theta$, followed by a temperature-scaled prediction layer, defined by $h_{\phi,\tau}:\mathcal{F} \to [0,1]^{Y}$ for classification
and $h_{\phi,\tau}:\mathcal{F} \to [0,1]^{W \times H \times Y}$ for segmentation, and parameterized by weights $\phi \in \Phi$ and temperature $\tau \in \mathcal{T} \subset \mathbb{R}$.
This temperature modifies the final softmax function as follows 
$\mathsf{softmax}(l(y),\tau) = \exp(l(y)/\tau)/\sum_{y'}\exp(l(y'),\tau)$, where $l(y)$ denotes the logit for class $y$ -- when omitted in the following equations, assume $\tau=1$. 
The backbone model and classification layer are trained using the Fair Identity Scaling (FIS) loss function~\cite{luo2024fairvisionequitabledeeplearning}, which is designed to balance individual and group-level fairness, with
\begin{equation}
\scalebox{0.9}{$
\theta^*,\phi^* = \arg\min\limits_{\theta,\phi} \frac{1}{N}  \sum\limits_{ (x,y,a) \in \mathcal{D}} \mathsf{w}(x,y,a,c) \times \ell \left (h_{\phi} \left ( f_{\theta}(x)\right) ,y  \right ),
$}
\label{eq:fis_optimization}
\end{equation}
where \( \ell(h_{\phi} \left ( f_{\theta}(x)\right), y) \) is the classification loss and $\mathsf{w}(x,y,a,c) = \left ( (1-c) \times \mathsf{s}^{I}(x,y) + c \times \mathsf{s}^{G}(a) ) \right )$, with  $\mathsf{s}^{I}(x,y) = \frac{\exp(\ell(h_{\phi} \left ( f_{\theta}(x)\right), y))}{\sum\limits_{ (\tilde{x},\tilde{y},\tilde{a}) \in \mathcal{D}} \exp(\ell(h_{\phi} \left ( f_{\theta}(\tilde{x})\right), \tilde{y}))}$ representing the individual scaling weight that gives higher weight to samples that are challenging for the model, 
%$\mathsf{s}^{G}(x,y,a) =  \frac{\exp\left(\mathsf{o}\left(\{\ell(h_{\phi} \left ( f_{\theta}(x)\right), y)\}_{(x,y,a) \in \mathcal{D}}, \{\ell(h_{\phi} \left ( f_{\theta}(\tilde{x})\right), \tilde{y})\}_{(\tilde{x},\tilde{y},\tilde{a}) \in \mathcal{D}_{a}}\right)\right)}{\sum_{g \in \mathcal{A}} \exp\left(\mathsf{o}\left(\{\ell(h_{\phi} \left ( f_{\theta}(x)\right), y)\}_{(x,y,a) \in \mathcal{D}},  \{\ell(h_{\phi} \left ( f_{\theta}(\tilde{x})\right), \tilde{y})\}_{(\tilde{x},\tilde{y},\tilde{a}) \in \mathcal{D}_{a}}\right)\right)}$ 
$\mathsf{s}^{G}(a) =  \frac{\exp\left(\mathsf{o}(\mathcal{L},\mathcal{L}_a) \right)} {\sum_{g \in \mathcal{A}} \exp\left(\mathsf{o}(\mathcal{L},\mathcal{L}_g) \right)}$
denoting the group scaling weight that promotes balanced learning across groups by computing the optimal transport function $\mathsf{o}(\mathcal{L},\mathcal{L}_g)$ that measures the distance between group loss distribution $\mathcal{L}_g = \{\ell(h_{\phi} \left ( f_{\theta}(x)\right), y)\}_{(x,y,a) \in \mathcal{D}|a=g}$ and the overall loss distribution $\mathcal{L} = \{\ell(h_{\phi} \left ( f_{\theta}(x)\right), y)\}_{(x,y,a) \in \mathcal{D}}$,
and $c \in [0,1]$ being a hyper-parameter controlling the balance between individual and group scaling weights.
The value of $c$ controls whether training prioritizes overall accuracy ($c \approx 0$) or fairness ($c \approx 1$) -- for this pre-training, we set $c=0.5$ for balanced training.
The loss function in~\eqref{eq:fis_optimization} automatically adjusts the sample weight $\mathsf{w}(.)$ based on sample-specific and group-level losses to ensure that the backbone model learns an overall accurate and fair feature space produced by the backbone feature extractor $f_{\theta^{*}}(.)$.

\subsection{Teacher Model Training (Step 1)}
\label{sec:training_step_1}

The cohort-specific teacher models are formed by adding a new randomly initialized classification layer $h_{\phi^{(t)}_{g}}(.)$, for each of the teachers $g \in \{0,\cdots,A\}$, to the pre-trained backbone model $f_{\theta^{*}}(.)$ from~\eqref{eq:fis_optimization}, and fine tune each model by greedily minimizing classification risk for each demographic cohort $g\in \{0,\cdots,A\}$, with the following loss:
\begin{equation}
\scalebox{0.85}{$
    \phi^{(t)^*}_{g}= \arg\min\limits_{\phi^{(t)}_{g}} \frac{1}{|\mathcal{D}_g|} \sum\limits_{(x,y,a) \in \mathcal{D}_g} \mathsf{w}(x,y,a,c) \times \ell \left(h_{\phi_{g}^{(t)}} \left (f_{\theta^{*}}(x),y \right ) \right),
$}
    \label{eq:optimization_teacher}
\end{equation}
where \( \ell(h_{\phi} \left ( f_{\theta}(x)\right), y) \) is the classification loss, $\mathsf{w}(x,y,a,c)$ is the sample weight defined in~\eqref{eq:fis_optimization} with $c=0$ to focus exclusively on the greedy individual scaling,  
$\mathcal{D}_g = \{(x,y,a) | (x,y,a) \in \mathcal{D}, a=g\}$, and $|.|$ is the set cardinality operator.
Note that in~\eqref{eq:optimization_teacher}, we only optimize the classification layer $h_{\phi_{g}^{(t)}} (.)$, leaving the backbone model $f_{\theta^{*}}(.)$ frozen.
This setup allows each teacher model to achieve high accuracy for its own demographic group, which is an essential step for the knowledge distillation detailed in the next section.

\subsection{Student Model Training (Step 2)}
\label{sec:training_step_2}

In the final step, we train a unified student model by introducing a randomly initialized classification layer, denoted  $h_{\phi^{(s)}}(\cdot)$ , onto the pre-trained backbone model $f_{\theta^{*}}(\cdot)$ from Step 0 in~\eqref{eq:fis_optimization}. This training minimizes the following loss function:
\begin{equation}
\scalebox{0.85}{$
\begin{aligned}[b]
    \phi^{(s)^*}= & \arg  \min\limits_{\phi^{(s)}} \frac{1}{N} \\ & \sum\limits_{(x,y,a) \in \mathcal{D}} \lambda \times \tau^2 \times \mathbb{KL}\left[ h_{\phi^{(s)}} \left (f_{\theta^{*}}(x) \right ) || h_{\phi^{(t)^*}_{a},\tau} \left (f_{\theta^{*}}(x) \right ) \right] + \\ 
    & \;\;\;\;\;\;\;\; (1-\lambda) \times \mathsf{w}(x,y,a,c) \times \ell \left(h_{\phi^{(s)}} \left (f_{\theta^{*}}(x),y \right ) \right),
\end{aligned}
$}
    \label{eq:optimization_student}
\end{equation}
where $\mathbb{KL}\left[ . \right]$ measures the Kullbeck-Leibler divergence between the student's and group-specific teachers' predictions~\cite{hinton2015distilling}, $\lambda \in [0,1]$ is a hyper-parameter to balance knowledge distillation from the group-specific models and the FIS loss, and $\tau$ is the temperature scaling to soften the teacher logits~\cite{li2022asymmetric}.
In the training of~\eqref{eq:optimization_student}, we set $\lambda=0.95$, indicating that we weight more the knowledge distillation aspect of the optimization,  and $c=0.5$, suggesting a balanced FIS training between accuracy and fairness.
By combining the knowledge distillation from the teacher models with the balanced fairness-accuracy FIS loss, the student model is optimized to enhance the Pareto front achieved by FIS. This approach ensures high overall accuracy, strong group-specific performance, and fair classification across groups.

\section{Experiment}
\label{sec:exp}

We evaluate FairDi on classification and segmentation tasks to showcase its versatility. This section outlines the experimental setup, datasets, evaluation metrics, competing methods, and statistical tests, followed by classification and segmentation results and an ablation study. Finally, we compare training and testing times.

\subsection{Experimental Setup}
\label{sec:experimental_setup}

\textbf{Training:} The classification experiments use the ResNet18~\cite{he2016deep} backbone. 
%, which is configured differently for the backbone, teacher, and student models.
In Step 0, we use  
%For the backbone, the ResNet18 model is trained with the entire training set using the 
Adam optimizer, a weight decay of $1\times10^{-4}$, and an initial learning rate of $1\times10^{-4}$ that is decayed by a factor of $0.1$ every $10$ epochs, for a total of $30$ epochs. 
To prevent overfitting, we apply an early stopping criterion where training halts if the validation worst-case area under the receiver operating characteristic curve (AUC) does not improve over $5$ consecutive epochs. 
%During the backbone training, the fusion weight $c=0.5$ in the FIS loss function is used to balance between individual and group scaling, allowing the model to capture both challenging individual cases and address group fairness.
For Step 1, 
%As explained in Sec.~\ref{sec:training_step_1}, 
the teacher models are initialized from this backbone, with the final classification layer being randomly initialized. For training the teacher models, we use the SGD optimizer with a momentum of $0.9$.   
%and configure the FIS loss with a fusion weight of $c = 0$, focusing exclusively on individual scaling without group scaling. 
%This training allows each teacher model to adapt specifically to its cohort, without deviating too much from the original backbone model.
Early stopping is applied here as well, halting training if the respective cohort validation AUC does not improve over $5$ consecutive epochs.
For Step 2, the student model is initialized from the backbone model with a randomly initialized classification layer, and is trained using an SGD optimizer with a momentum of $0.9$.
%For the student model, we employ a combined knowledge distillation loss, which includes a KL-divergence-based distillation loss and a ground truth loss calculated using FIS. 
The distillation loss is configured with a temperature value of \textcolor{blue}{$\tau=1.5$} in \cref{eq:optimization_student} to soften the teacher logits. 
%, and an alpha weight of $0.95$ to balance the distillation loss with the ground truth loss. 
%The FIS fusion weight is set to $c = 0.5$ during student model training to support both cohort-specific adaptation and fairness-driven generalization. 
Early stopping is applied to the student model based on the validation worst-case AUC, with training terminating if no improvement is observed over $5$ epochs.  
We use CutMix~\cite{yun2019cutmix} to augment the training data for the training of backbone, teacher and student models.
Our experiments are conducted on a computer using one NVIDIA RTX A6000 GPU. The implementation is built on Python $3.12.7$ and PyTorch $2.2.2$.

The segmentation experiments utilize the TransUNet backbone, built on the ViT-B (Vision Transformer) architecture, tailored specifically for medical imaging. The model is trained using the AdamW optimizer with a learning rate of 0.01, momentum of 0.9, weight decay of \(1 \times 10^{-4}\), and exponential learning rate decay. In the initial stages of training, a warmup strategy is employed to stabilize model training with the Harvard-FairSeg dataset~\cite{tian2024fairseg}. The TransUNet model is trained for 300 epochs without early stopping, following its original implementation, ensuring robust segmentation performance. The batch size for training is set to 42. The warmup strategy is applied at the beginning of training to stabilize the models.

\textbf{Datasets:} For the classification task, we conduct our experiments on five medical imaging datasets that have been used to benchmark fairness methods~\cite{zong2022medfair}--they are: Fitzpatrick17K~\citep{groh2021evaluating}, HAM10000~\cite{tschandl2018ham10000}, Papila~\cite{kovalyk2022papila}, CheXpert~\cite{irvin2019chexpert}, and MIMIC-CXR~\cite{johnson2019mimic}. 
%These datasets cover a diverse range of medical tasks in various anatomical regions, a large range of sizes, and they have been used by many fairness methods to assess the ability of these methods to produce unbiased classification results in terms of sensitive attributes such as skin type, age, sex, and race. 
To ensure consistent data preparation across all datasets, we follow a standardized pre-processing approach inspired by~\cite{zong2022medfair}. For example, we binarize sensitive attributes (e.g., skin type, age, sex, and race) and classification labels to facilitate evaluation of demographic fairness. 
More specifically, HAM10000 has classes benign and malignant, and two sensitive attributes: ‘Age’ (binarized into 0 to 60 and above 60) and ‘Gender’ (Male and Female). Fitzpatrick17k has classes malignant and non-malignant, with the ‘Skin Type’ attribute binarized into one group with skin types ranging from 0 to 2, and another group with skin types above 2. The PAPILA dataset has classes healthy and glaucoma, with two sensitive attributes: ‘Age’ (binarized into 0 to 60 and above 60) and ‘Gender’ (Male and Female). The CheXpert and MIMIC-CXR has classes 'No-Finding' and 'Any Finding', with three sensitive attributes: ‘Age’ (binarized into 0 to 60 and above 60), ‘Gender’ (Male and Female), and ‘Race’ (Non-White, including Black or Asian patients, and White).
Additionally, as instructed by the MedFair benchmark~\cite{zong2022medfair}, samples with incomplete or missing data are excluded from the analysis. 
%\textcolor{red}{What does it mean to remove studies with incomplete or missing data? Do other methods do the same (e.g., MedFair)? (Removing studies with incomplete or missing data typically means excluding samples where essential information, like sensitive attributes, is not recorded. In this case, we remove samples that lack the sensitive attribute (e.g., data with NaN values for this attribute), as including them could lead to biased or incomplete analysis. Yes, MedFair and other similar studies also follow this approach. They exclude samples without recorded sensitive attributes to ensure that analyses are conducted on a complete dataset where demographic or subgroup information is available, supporting a more accurate and fair model evaluation.)}
%These pre-processing steps aim to create uniform conditions for evaluating fairness while accounting for potential biases in the source data. 
A more detailed description of the pre-processing is provided in Sec.~\ref{sec:data_preprocessing} of the supplementary material.

For the segmentation task, we use the Harvard-FairSeg dataset~\cite{tian2024fairseg}, which is a scanning laser ophthalmoscopy (SLO) fundus image dataset with a disc-cup segmentation task to assess the optic nerve head structures and diagnosing glaucoma in its early stage. This Harvard-FairSeg dataset has 10,000 samples collected from 10,000 unique subjects. We split the data into an 8,000 for training and a 2,000 for testing. The subjects in the dataset have a mean age of 60.3 years, with a standard deviation of 16.5 years. Six sensitive attributes—age, gender, race, ethnicity, preferred language, and marital status—are available to support fairness-focused studies. In this paper, we focus on three sensitive attributes with the highest observed group-wise discrepancies: race, gender, and ethnicity.

More details about the classification and segmentation datasets are shown in~\cref{sec:dataset_details} in the supp. material.

\textbf{Evaluation Measures and Statistical Tests:} Binary classification is assessed with AUC. 
To evaluate fairness, we report metrics that account for performance disparities across demographics cohorts. 
For instance, AUC Gap~\cite{zong2022medfair} measures the performance difference between the highest and lowest-performing cohorts. 
The Worst-Case AUC~\cite{zong2022medfair}  calculates the minimum AUC achieved among all cohorts. 
Another fairness measure that we present is the Equity-Scaled AUC (ES-AUC)~\cite{luo2024fairvisionequitabledeeplearning}, defined in Sec.~\ref{sec:fairness_measures} of the supplementary material, which adjusts the overall AUC by penalizing discrepancies between the AUCs of the cohorts. 
We also measure the Performance-Scaled Disparity (PSD) that evaluates fairness by comparing cohort performance relative to overall AUC--please see Sec.~\ref{sec:fairness_measures} of the supplementary material for the definition of the MeanPSD, which calculates the standard deviation of AUC scores across cohorts, and MaxPSD that calculates the maximum absolute difference between cohorts' AUCs. 

We evaluate the segmentation tasks using the Dice Similarity Coefficient~\cite{dice1945measures,sorensen1948method}, which measures the overlap between predicted and ground truth segmentations, and the  
%, calculated as \( \mathsf{Dice} = \frac{2 \times |A \cap B|}{|A| + |B|} \), where \(A\) and \(B\) represent the predicted and actual segmented pixels, respectively. 
Intersection over Union (IoU)~\cite{jaccard1901etude} that measures the ratio of the intersection to the union of the predicted and ground truth segments. 
%, given by \( \text{IoU} = \frac{|A \cap B|}{|A \cup B|} \). Both metrics range from 0 to 1, with 1 indicating perfect overlap, but Dice tends to be slightly more sensitive to small overlaps than IoU due to its formula.
In the FairSeg benchmark~\cite{tian2024fairseg}, 
Equity-scaled Dice (ES-Dice) and IoU (ES-IoU) modify the Dice and IoU to penalize discrepancies in performance across demographic groups by scaling the overlap scores down if subgroup performance is uneven. In essence, equity-scaled metrics reduce the overall score when there are larger performance gaps between cohorts, encouraging models to achieve consistent segmentation quality across all subgroups--please see Sec.~\ref{sec:fairness_measures} of the supplementary material for the definition of ES-Dice and ES-IoU.

%\textcolor{red}{Evaluation measures of segmentation task, we ???? TAHIR and MILAD, please add the details about the evaluation of the segmentation task} 

As suggested in~\cite{zong2022medfair}, we test significance by applying the Friedman test~\cite{friedman1937use}, followed by the Nemenyi post-hoc test~\cite{nemenyi1963distribution}, to detect significant performance differences between algorithms. If the Friedman test shows significance (p-value $< 0.05$), the Nemenyi test averages the ranks of each algorithm across datasets. The results are displayed using Critical Difference (CD) diagrams, where algorithms connected by a line are statistically similar, while those in separate groups are significantly different.

\textbf{Baselines and State-of-the-art (SOTA) Methods:} We compare our proposed FairDi against several baseline and SOTA fairness methods.
For the classification task, a common baseline in fairness studies is the Empirical Risk Minimization (ERM)~\cite{vapnik1999overview}, which is trained by minimizing the average classification error across the training set without accounting for demographic cohorts or sensitive attributes. 
While ERM optimizes overall accuracy, it overlooks group-specific biases, making it a valuable baseline for assessing fairness gaps. 
GroupDRO~\cite{sagawa2019distributionally} is designed to minimize the worst-case loss across demographic groups. 
By focusing on the group with the highest training loss, GroupDRO emphasizes performance fairness, aiming to protect disadvantaged cohorts. 
Another baseline, SWAD~\cite{cha2021swad}, explores a range of flat minima in the loss landscape through dense sampling of model weights. This approach is particularly beneficial for enhancing model generalization across diverse settings, but does not directly address fairness across demographic groups. 
%\textcolor{red}{The Learning Not to Learn (LNL) method~\cite{kim2019learning} reduces bias iteratively by minimizing the mutual information between the feature representations and the bias. The Entangle and Disentangle (EnD) method~\cite{tartaglione2021end} separates confounding factors by implementing an "information bottleneck," allowing only essential information to pass through. Orthogonal Disentangled Representations (ODR)~\cite{sarhan2020fairness} achieves separation of useful and sensitive representations by applying orthogonality constraints to ensure independence between them. Sharpness-Aware Minimization (SAM)~\cite{foret2020sharpness} optimizes parameters to locate regions in the parameter space with consistently low loss, enhancing model robustness.} 
We also compare our approach  to FIS~\cite{luo2024fairvisionequitabledeeplearning}, the SOTA fairness method that applies a scaling mechanism to adjust loss contributions based on demographic cohort performance. 

We evaluate fair segmentation with the TransUNet backbone~\cite{chen2021transunet}, which combines CNNs with transformers, leveraging self-attention for capturing long-range dependencies while preserving spatial hierarchies, making it effective for medical image segmentation.
The first TransUNet-based fair segmentation method is implemented by combining it with the Adversarially Fair Representations (ADV)~\cite{madras2018learning} to create unbiased representations by making it difficult to infer sensitive attributes, thus reducing bias. 
Another fair TransUNet method is achieved by combining it with GroupDRO~\cite{sagawa2019distributionally}, which improves model fairness by minimizing the maximum training loss across groups, using regularization to prevent bias toward any single group.
We also compare with the SOTA TransUNet+Fair Error-Bound Scaling (FEBS)~\cite{tian2024fairseg} that improves fairness by rescaling the loss based on each identity group’s upper training error bound.

%The methods above provide a comprehensive set of fairness methods to assess the performance of our FairDi method both in classification and segmentation tasks. 
%\textcolor{red}{We may need to introduce more methods here, depending on the CD diagrams. (Done)}

%%
\subsection{Classification Results}

% Start of the table
\begin{table}[t!]
    \centering
    \caption{
    %\textcolor{red}{We should update this table based on adding new STOA methods.} 
    Summary of the evaluation and ranking of fairness models (ERM, GroupDRO, SWAD, FIS, and our FairDi) across \textbf{classification} benchmarks HAM10000 (attributes: Age, Gender), Fitzpatrick17k (attribute: Skin Type), PAPILA (attributes: Age, Gender), CheXpert (attributes: Age, Gender, Race), and MIMIC-CXR (attributes: Age, Gender, Race). We report overall AUC, minimum AUC, ES-AUC, AUC gap, MeanPSD, and MaxPSD. Best results are highlighted and detailed results are shown in Table~\ref{tab:comparison_results_complete} of the supplementary material. 
    \label{tab:comparison_results_summary}}
 
    \scalebox{0.64}{
    \begin{tabular}{l c c c c c c c}

\toprule
\multicolumn{3}{c}{\textbf{Measures and Ranks}} & \textbf{ERM} & \textbf{GroupDRO} & \textbf{SWAD} & \textbf{FIS} & \textbf{FairDi (Ours)} \\ 
\midrule
% \endhead

% \midrule \multicolumn{8}{r}{{Continued on next page}} \\ \midrule
% \endfoot

% \bottomrule
% \endlastfoot

% Summary row
\multicolumn{3}{c}{Avg. Overall AUC Score$\uparrow$} & 0.8485 & 0.8511 & 0.8735 & 0.8976 & \textbf{0.9137} \\ 
\multicolumn{3}{c}{Avg. Min. AUC Score$\uparrow$} & 0.8082 & 0.8127 & 0.8350 & 0.8815 & \textbf{0.9050} \\ 
\multicolumn{3}{c}{Avg. ES-AUC Score$\uparrow$} & 0.8264 & 0.8253 & 0.8474 & 0.8864 & \textbf{0.9062} \\ 
\multicolumn{3}{c}{Avg. AUC Gap Score$\downarrow$} & 0.0589 & 0.0710 & 0.0683 & 0.0249 & \textbf{0.0166} \\ 
\multicolumn{3}{c}{Avg. MeanPSD Score$\downarrow$} & 0.0379 & 0.0440 & 0.0410 & 0.0141 & \textbf{0.0091} \\ 
\multicolumn{3}{c}{Avg. MaxPSD Score$\downarrow$} & 0.0758 & 0.0881 & 0.0820 & 0.0282 & \textbf{0.0182} \\ 
\midrule

% Ranking row
\multicolumn{3}{c}{Avg. Overall AUC Rank$\downarrow$} & 3.55 & 4.45 & 2.55 & 2.91 & \textbf{1.55} \\ 
\multicolumn{3}{c}{Avg. Min. AUC Rank$\downarrow$} & 3.91 & 4.36 & 2.64 & 2.73 & \textbf{1.36} \\ 
\multicolumn{3}{c}{Avg. ES-AUC Rank$\downarrow$} & 3.82 & 4.45 & 2.45 & 2.73 & \textbf{1.55} \\ 
\multicolumn{3}{c}{Avg. AUC Gap Rank$\downarrow$} & 3.00 & 3.55 & 3.27 & 3.00 & \textbf{2.18} \\ 
\multicolumn{3}{c}{Avg. MeanPSD Rank$\downarrow$} & 3.00 & 3.55 & 3.27 & 3.00 & \textbf{2.18} \\ 
\multicolumn{3}{c}{Avg. MaxPSD Rank$\downarrow$} & 3.00 & 3.55 & 3.27 & 3.00 & \textbf{2.18} \\ 
\bottomrule

\end{tabular}
}
\end{table}

% Start of the table
\begin{table}[t!]
    \centering
    \caption{Summary of the evaluation and ranking of fairness models (TransUNet, TransUNet+ADV, TransUNet+GroupDRO, TransUNet+FEBS, and our TransUNet+FairDi)  across the \textbf{segmentation} of optic cup and rim on the Harvard-FairSeg dataset, with race, gender and ethnicity sensitive attributes. We report Overall ES-Dice, Overall Dice, Overall ES-IoU, and Overall IoU. Best results are highlighted and detail results are shown in~\cref{tab:fairseg_race,tab:fairseg_gender,tab:fairseg_ethnicity} at the supplementary material.     \label{tab:comparison_results_summary+segmentation}}
 
    \scalebox{0.56}{
    \begin{tabular}{l c c c c c c c}

\toprule
\multicolumn{3}{c}{\textbf{Measures and Ranks}} & \textbf{TransUNet} & \textbf{TransUNet} & \textbf{TransUNet} & \textbf{TransUNet} & \textbf{TransUNet} \\ 
& & & & \textbf{+} & \textbf{+} & \textbf{+} & \textbf{+} \\ 
& & & & \textbf{ADV} & \textbf{GroupDRO} & \textbf{FEBS} & \textbf{FairDi} \\ 
& & & &  &  &  & \textbf{(Ours)} \\ 
\midrule

% Summary row
\multicolumn{3}{c}{Avg. Overall ES-Dice Score$\uparrow$} & 0.7955 & 0.7873 & 0.7956 & 0.7966 & \textbf{0.8062} \\ 
\multicolumn{3}{c}{Avg. Overall Dice Score$\uparrow$} & 0.8204 & 0.8112 & 0.8193 & 0.8204 & \textbf{0.8260} \\ 
\multicolumn{3}{c}{Avg. Overall ES-IoU Score$\uparrow$} & 0.6871 & 0.6767 & 0.6870 & 0.6889 & \textbf{0.7002} \\ 
\multicolumn{3}{c}{Avg. Overall IoU Score$\uparrow$} & 0.7119 & 0.7003 & 0.7106 & 0.7124 & \textbf{0.7209} \\ 
\midrule

% Ranking row
\multicolumn{3}{c}{Avg. Overall ES-Dice Rank$\downarrow$} & 3.17 & 4.67 & 3.00 & 3.00 & \textbf{1.17} \\ 
\multicolumn{3}{c}{Avg. Overall Dice Rank$\downarrow$} & 3.00 & 4.83 & 3.50 & 2.67 & \textbf{1.00} \\ 
\multicolumn{3}{c}{Avg. Overall ES-IoU Rank$\downarrow$} & 3.17 & 4.50 & 3.67 & 2.67 & \textbf{1.00} \\ 
\multicolumn{3}{c}{Avg. Overall IoU Rank$\downarrow$} & 2.67 & 4.83 & 3.83 & 2.67 & \textbf{1.00} \\ 
\bottomrule

\end{tabular}
}
\end{table}

\begin{figure*}[t!]
  \centering
  \includegraphics[width=\linewidth]{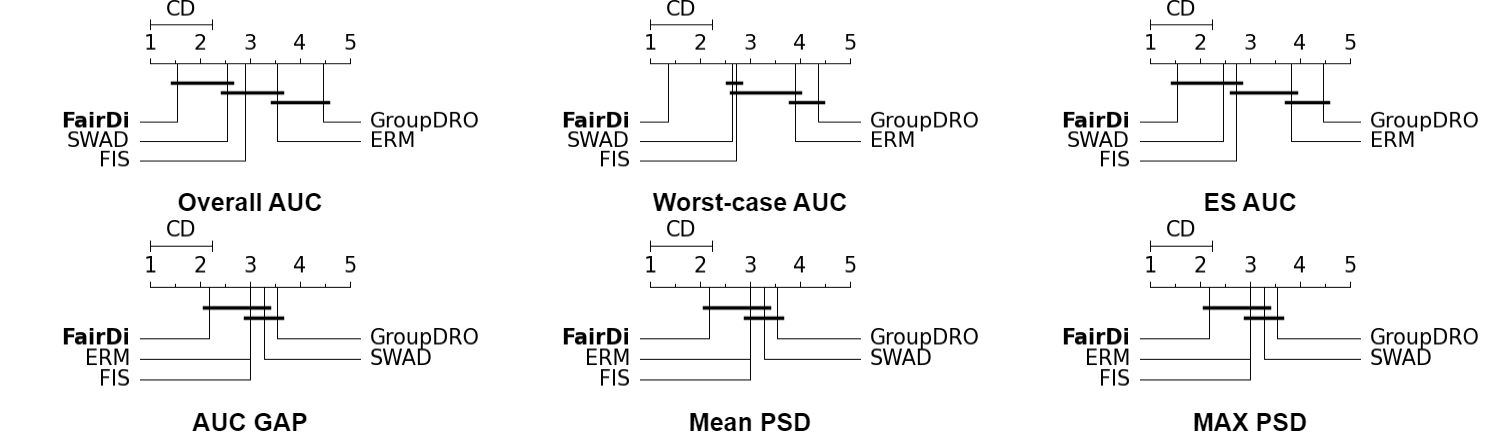}
   \caption{Performance of fairness algorithms for classification  across all datasets as average rank CD diagrams. Our FairDi is the highest ranked method for all settings, being significantly better than most methods, and for the worst-case AUC, it is the single best method.}
   \label{fig:CD Classification}
\end{figure*}

%This section provides a comparative analysis of our FairDi method against baseline and state-of-the-art (SOTA) methods across multiple datasets, using various metrics: overall AUC, subgroup-minimum AUC, ES-AUC, AUC gap, MeanPSD, and MaxPSD.
Table~\ref{tab:comparison_results_summary} summarizes results across all datasets and sensitive attributes, where complete results are available in Table~\ref{tab:comparison_results_complete} of the supplementary material. FairDi achieves the highest average metrics across all measures, with an Average Overall AUC of 0.9137, outperforming FIS (0.8976), SWAD (0.8735), GroupDRO (0.8511), and ERM (0.8485), demonstrating superior robustness and consistency across sensitive attributes.
In fairness metrics, FairDi attains the highest Average Minimum AUC Score (0.9050), indicating robustness across the most challenging subgroup, compared to FIS (0.8815) and others. FairDi also leads in Average ES-AUC Score at 0.9062, outperforming FIS (0.8864), SWAD (0.8474), GroupDRO (0.8253), and ERM (0.8264).
FairDi achieves the lowest Average AUC Gap Score of 0.0166, indicating minimal performance discrepancy across groups, while FIS follows with 0.0249. SWAD, GroupDRO, and ERM show significantly higher AUC Gaps, highlighting FairDi’s effectiveness in reducing group disparities. Similarly, FairDi achieves the lowest MeanPSD (0.0091) and MaxPSD (0.0182) scores, reflecting its ability to provide fairer outcomes across demographics.

%In overall rankings, FairDi consistently ranks highest across fairness and accuracy, confirming its robust improvement over other methods. 
\cref{fig:CD Classification} shows the Nemenyi post-hoc test results on Overal AUC, Worst-case AUC, ES-AUC, AUC Gap, Mean PSD and Max PSD settings with raw data in \cref{tab:comparison_results_complete} of the supplementary material.
Our FairDi is the highest ranked method for all settings, where for most settings, it is significantly better than between 1 to all four competing methods.
It is worth noting that for the worst-case AUC, our method is significantly better than all others.
These results demonstrate FairDi’s ability to enhance both classification accuracy and fairness, significantly reducing performance gaps across demographic subgroups.

\subsection{Segmentation Results}

% \begin{table}[t!]
%     \centering
%     \caption{
%     Summary of segmentation performance across methods evaluated on Harvard-FairSeg with overall ES-Dice, Dice, ES-IoU, and IoU scores, and their respective ranks. Best results are highlighted.
%     \label{tab:segmentation_performance_summary}}
 
%     \scalebox{0.4}{
%     \begin{tabular}{l c c c c c}
%     \toprule
%     \textbf{Measures and Ranks} & \textbf{TransUNet} & \textbf{TransUNet + ADV} & \textbf{TransUNet + GroupDRO} & \textbf{TransUNet + FEBS} & \textbf{TransUNet + FairDi} \\ 
%     \midrule

%     Avg. Overall ES-Dice Score & 0.7955 & 0.7873 & 0.7956 & 0.7966 & \textbf{0.8062} \\ 
%     Avg. Overall Dice Score & 0.8204 & 0.8112 & 0.8193 & 0.8204 & \textbf{0.8260} \\ 
%     Avg. Overall ES-IoU Score & 0.6871 & 0.6767 & 0.6870 & 0.6889 & \textbf{0.7002} \\ 
%     Avg. Overall IoU Score & 0.7119 & 0.7003 & 0.7106 & 0.7124 & \textbf{0.7209} \\ 
%     \midrule

%     Avg. Overall ES-Dice Rank & 3.17 & 4.67 & 3.00 & 3.00 & \textbf{1.17} \\ 
%     Avg. Overall Dice Rank & 3.00 & 4.83 & 3.50 & 2.67 & \textbf{1.00} \\ 
%     Avg. Overall ES-IoU Rank & 3.17 & 4.50 & 3.67 & 2.67 & \textbf{1.00} \\ 
%     Avg. Overall IoU Rank & 2.67 & 4.83 & 3.83 & 2.67 & \textbf{1.00} \\ 
%     \bottomrule
%     \end{tabular}
%     }
% \end{table}

\begin{figure}[t]
  \centering
  \includegraphics[width=\linewidth]{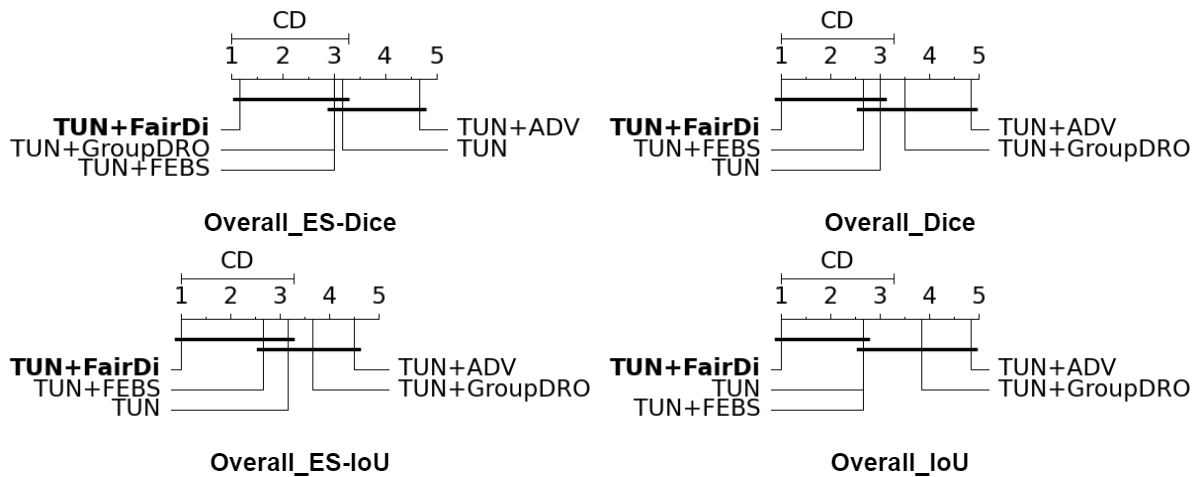}
   \caption{Performance of fairness segmentation algorithms shown with average rank CD diagrams. FairDi consistently outperforms most methods, ranking the highest in all settings. In this figure, TUN stands for TransUNet}
   \label{fig:CD_Segmentation}
\end{figure}

%Using the TransUNet~\cite{chen2021transunet} backbone, we compare our proposed TransUNet+FairDi to baselines TransUNet+ADV~\cite{madras2018learning} and TransUNet+GroupDRO~\cite{sagawa2019distributionally}, and SOTA TransUNet+FEBS~\cite{tian2024fairseg}, using the measures Overall Dice, Overall ES-Dice, Overall IoU, Overall ES-IoU on the segmentation of optic cup and rim on the Harvard-FairSeg dataset~\cite{tian2024fairseg}.
Table~\ref{tab:comparison_results_summary+segmentation} shows a summary of the results of TransUNet~\cite{chen2021transunet}, TransUNet+ADV~\cite{madras2018learning},  TransUNet+GroupDRO~\cite{sagawa2019distributionally}, TransUNet+FEBS~\cite{tian2024fairseg}, and our proposed TransUNet+FairDi across race, gender and ethnicity sensitive attributes, where the complete results are available in~\cref{tab:fairseg_race,tab:fairseg_gender,tab:fairseg_ethnicity} of the supplementary material.
FairDi achieves the highest average metrics across all measures, with an Average Overall Dice of 0.8260, which is superior to FEBS (0.8204), GroupDRO (0.8193), ADV (0.8112), and TransUNet (0.8204). Regarding fairness metrics, FairDi achieves an Overall ES-Dice of 0.8062, outperforming FEBS (0.7966), GroupDRO (0.7956), ADV (0.7873), and TransUNet (0.7955). Similar results are observed for IoU and ES-IoU, demonstrating the ability of our approach achieve fairer segmentation across various cohorts.

%Regarding the rankings, FairDi is consistently ranked highest in terms of fairness and accuracy, confirming its robust improvement over other methods. 
The Nemenyi post-hoc test results on Overall Dice, Overall ES-Dice, Overall IoU, Overall ES-IoU shown in~\cref{fig:CD_Segmentation}, with raw data in~\cref{tab:fairseg_race,tab:fairseg_gender,tab:fairseg_ethnicity} at the supplementary material, demonstrate that our FairDi is the highest ranked method for all settings. Note that for most settings, it is significantly better than between 1 and 2 competing methods.
These results confirm FairDi’s ability to improve segmentation accuracy and fairness.

\subsection{Ablation Study}

\begin{table*}[t!]
    \centering
    \caption{Ablation study on HAM10000 (with attribute Gender) and PAPILA (with attribute Gender). Starting from ERM, followed by ERM + CutMix, FIS, FIS + CutMix (Step 0), FairDi-Teacher (Female - Step 1), FairDi-Teacher (Male - Step 1), and FairDi-Student (Step 2). \label{tab:ablation}}
    \scalebox{0.55}{
    \begin{tabular}{l p{3.10cm} p{3.10cm} p{3.10cm} p{3.10cm} p{3.10cm} p{3.10cm} p{3.10cm} p{3.10cm}}
    \toprule
    % \textbf{Dataset (Attribute)} & \textbf{Metric} & \textbf{ResNet18 + BCEWithLogits Loss} & \textbf{ResNet18 + BCEWithLogits Loss + CutMix} & \textbf{ResNet18 + FISLoss} & \textbf{ResNet18 + FISLoss + CutMix} & \textbf{ResNet18 + FISLoss + CutMix + Teacher Female} & \textbf{ResNet18 + FISLoss + CutMix + Teacher Male} & \textbf{ResNet18 + FISLoss + Distillation Loss + CutMix} \\ 
    \textbf{Dataset (Attribute)} & \textbf{Metric} & \textbf{ERM} & \textbf{ERM + CutMix} & \textbf{FIS} & \textbf{FIS + CutMix \newline (Step 0)} & \textbf{FairDi-Teacher \newline (Female - Step 1)} & \textbf{FairDi-Teacher \newline (Male - Step 1)} & \textbf{FairDi-Student \newline (Step 2)} \\ 
    \midrule

    % HAM10000 data
    \multirow{7}{*}{HAM10000 (Gender)} 
    & Overall AUC↑ & 0.8890 & 0.8952 & 0.9174 & 0.9300 & 0.9420 & 0.9502 & \textbf{0.9560} \\ 
    & Female AUC↑ & 0.8826 & 0.8785 & 0.9151 & 0.9276 & \textbf{0.9585} & 0.9475 & 0.9554 \\ 
    & Male AUC↑ & 0.8956 & 0.9157 & 0.9222 & 0.9287 & 0.9300 & 0.9510 & \textbf{0.9553} \\ 
    & AUC Gap↓ & 0.0130 & 0.0372 & 0.0071 & 0.0011 & 0.0285 & 0.0035 & \textbf{0.000029} \\ 
    & ES-AUC↑ & 0.8833 & 0.8789 & 0.9142 & 0.9283 & 0.9288 & 0.9485 & \textbf{0.9553} \\ 
    & MeanPSD↓ & $7.31 \times 10^{-3}$ & $2.08 \times 10^{-2}$ & $3.87 \times 10^{-3}$ & $5.77 \times 10^{-4}$ & $1.51 \times 10^{-2}$ & $1.84 \times 10^{-3}$ & $\mathbf{1.52 \times 10^{-5}}$ \\ 
    & MaxPSD↓ & $1.46 \times 10^{-2}$ & $4.16 \times 10^{-2}$ & $7.74 \times 10^{-3}$ & $1.15 \times 10^{-3}$ & $3.03 \times 10^{-2}$ & $3.68 \times 10^{-3}$ & $\mathbf{3.03 \times 10^{-5}}$ \\ 
    \midrule

    % PAPILA data
    \multirow{7}{*}{PAPILA (Gender)} 
    & Overall AUC↑ & 0.8433 & 0.9142 & 0.8938 & 0.9391 & 0.9315 & 0.9395 & \textbf{0.9572} \\ 
    & Female AUC↑ & 0.8447 & \textbf{0.9714} & 0.8905 & 0.9471 & 0.9515 & 0.9027 & 0.9508 \\ 
    & Male AUC↑ & 0.8267 & 0.8214 & 0.9236 & 0.9186 & 0.8897 & 0.9841 & \textbf{0.9857} \\ 
    & AUC Gap↓ & \textbf{0.0180} & 0.1500 & 0.0331 & 0.0285 & 0.0618 & 0.0814 & 0.0349 \\ 
    & ES-AUC↑ & 0.8358 & 0.8504 & 0.8792 & 0.9259 & 0.9036 & 0.9028 & \textbf{0.9408} \\ 
    & MeanPSD↓ & $\mathbf{1.07 \times 10^{-2}}$ & $8.2 \times 10^{-2}$ & $1.85 \times 10^{-2}$ & $1.52 \times 10^{-2}$ & $3.32 \times 10^{-2}$ & $4.33 \times 10^{-2}$ & $1.82 \times 10^{-2}$ \\ 
    & MaxPSD↓ & $\mathbf{2.13 \times 10^{-2}}$ & $1.64 \times 10^{-1}$ & $3.70 \times 10^{-2}$ & $3.03 \times 10^{-2}$ & $6.63 \times 10^{-2}$ & $8.66 \times 10^{-2}$ & $3.64 \times 10^{-2}$ \\ 
    \bottomrule
    \end{tabular}
    }    
\end{table*}

%\textcolor{red}{In this section, we should report results same with other sections and tables without percentage.} 
In this ablation study, we explore the technical contributions of our FairDi model, on two classification datasets, to understand their impact on accuracy and fairness. 
Tab.~\ref{tab:ablation} presents the results across the HAM10000 and PAPILA datasets, focusing on Gender as the sensitive attribute. 
We start from the baseline ERM model (ResNet18) trained with binary cross entropy (BCE) loss~\cite{zong2022medfair}, which has moderate Overall AUC scores of $0.8890$ on HAM10000 and $0.8433$ on PAPILA. 
Integrating CutMix~\cite{yun2019cutmix}, the Overall AUC results improve to $0.8952$ on HAM10000 and $0.9142$ on PAPILA. 
However, this accuracy enhancement results in a larger AUC Gap, particularly on PAPILA, which reaches $0.1500$. 
%This suggests a trade-off where CutMix boosts accuracy at the cost of increased subgroup disparities. 
Replacing BCE with FIS loss~\cite{luo2024fairvisionequitabledeeplearning},  increases Overall AUC to $0.9174$ on HAM10000 and $0.8938$ on PAPILA and substantially reduces AUC Gap to $0.0071$ on HAM10000 and $0.0331$ on PAPILA, showing a better balance between accuracy and fairness. 
The combination of FIS Loss with CutMix leads to further improvements, reaching an Overall AUC of $0.9300$ on HAM10000 and $0.9391$ on PAPILA and further AUC Gap reductions to $0.0011$ on HAM10000 and $0.0285$ on PAPILA. This configuration achieves high ES-AUC and low MeanPSD and MaxPSD values, indicating a favorable balance between accuracy and fairness. This forms the backbone model from Step 0, explained in Sec.~\ref{sec:training_step_0}.

Step 1, explained in Sec.~\ref{sec:training_step_1}, consists of training group-specific FairDi-teacher models, which have high group-specific overall AUC scores, with Female AUC reaching $0.9585$ and Male AUC reaching $0.9510$ on HAM10000, and similar performances on PAPILA (Female AUC of $0.9515$ and Male AUC of $0.9841$). However, these biased models also introduce larger AUC Gap values ($0.0285$ on HAM10000 and $0.0814$ on PAPILA) due to the inherent performance discrepancy between the cohort-specific models. This suggests that while group-specific performances are strong, fairness is sacrificed.

The final knowledge distillation (from group-specific teacher models) and accuracy-fairness balanced training of the student model during Step 2, explained in Sec.~\ref{sec:training_step_2}, produces the FairDi-student model, which achieves an Overall AUC of $0.9560$ on HAM10000 and $0.9572$ on PAPILA. Both Female and Male AUCs remain high, while the model minimizes the AUC Gap to just $0.000029$ on HAM10000 and $0.0349$ on PAPILA. These results indicate a significant reduction in cohort-specific performance disparities compared to the biased teacher models, demonstrating the effectiveness of our FairDi design in balancing cohort-specific performance with overall accuracy and fairness.

\subsection{Training and Testing Running Times}

%\textcolor{red}{In this section, we compare the training and testing running times of our FairDi with ERM, SWAD, GroupDRO, and FIS. The running time for FairDi was found to be substantially higher than each of the other methods. Specifically,  FairDi required 148.3\% more time than ERM, 180.4\% more than SWAD, 168.9\% more than GroupDRO, and 165.2\% more than FIS. For example, when using the HAM10000 dataset with gender-sensitive attributes, the running times (training and testing combined) for ERM, SWAD, GroupDRO, and FIS were 197, 174.5, 182, and 184.5 seconds, respectively. In comparison, the FairDi method’s running time was structured across multiple training steps: Backbone Model Training (Step 0) took 184.5 seconds, Female and Male Teacher Models Training (Step 1) took 77.81 and 63.47 seconds, respectively, and Student Model Training (Step 2) took 163.53 seconds. Altogether, the total running time for the FairDi method amounted to 489.31 seconds. It is important to note that in the FairDi model, for both Teacher Models Training (Step 1) and Student Model Training (Step 2), we only train the final classifier. Despite this streamlined approach, which reduces training complexity, FairDi still exhibited a running time approximately 165.25\% longer than the other methods on average. This increase reflects the additional computational demands of FairDi, which may be justified by the performance improvements it provides.}

For the HAM10000 dataset with gender-sensitive attribute, the training times for ERM, SWAD, GroupDRO, and FIS were 19.75, 28.78, 19.12, and 17.56 minutes, respectively.
%, with a consistent testing time of 1.11 minutes for each. 
In comparison, the FairDi method’s total training time is structured across the following multiple steps: Backbone Model Training (Step 0) takes 17.56 minutes, Teacher (Female) Model Training (Step 1) takes 5.53 minutes, Teacher (Male) Model Training (Step 1) takes 4.26 minutes, and Student Model Training (Step 2) takes 16.96 minutes, resulting in a combined training time of $\approx 40$ minutes, assuming that the two teachers can be trained in parallel. 
The testing time for ERM, SWAD, GroupDRO, FIS, and FairDi is identical at 0.033 seconds per image. 
%1.11 minutes. \textcolor{red}{The testing time per image for these models is 0.033 seconds.} 
%It is important to note that in the FairDi model, for both Teacher Models Training (Step 1) and Student Model Training (Step 2), we only train the final classifier.
%\textcolor{red}{The backbone model contains a total of 11,177,025 parameters, all of which are trainable in the initial training phase. In the training of the female and male teacher models, $ \approx 0.5K $ parameters are trainable specifically in the classification layer, while $ \approx 11M $ parameters remain non-trainable, derived from the backbone model. Similarly, for the student model, $ \approx 0.5K $ parameters are allocated to the classification layer, with the same $ \approx 11M $ non-trainable parameters from the backbone model.}
In terms of memory footprint, the backbone model contains a total of $\approx$11M parameters, all of which are trainable at Step 0 of the training. The classification layers of the female and male teacher models introduce $\approx$0.5K (each) new parameters trainable at Step 1 of the training. Similarly, the student model adds another $\approx$0.5K new parameters at the classification layer, with the same $\approx$11M non-trainable parameters from the backbone model.

\section{Conclusion}
\label{sec:con}

In this work, we introduced FairDi, a novel fairness method that tackles the challenging trade-offs between accuracy, subgroup-specific performance, and fairness in medical imaging. Unlike traditional fairness approaches that attempt to balance these competing goals within a single model, FairDi leverages the strengths of biased teacher models trained specifically for sensitive subgroups. This unique strategy allows the unified student model to distill the teacher models' knowledge and achieve high overall accuracy, while reducing performance disparities across demographic groups.

Our extensive evaluations across multiple medical imaging datasets demonstrate that FairDi consistently outperforms current fairness methods, achieving both high accuracy and fairness. The experimental results highlight FairDi's ability to narrow the AUC gap between advantaged and disadvantaged subgroups, while also minimizing performance-scaled disparity (PSD), a crucial metric for real-world clinical relevance. Furthermore, by distilling knowledge from subgroup-optimized teachers, FairDi not only enhances the interpretability of fairness outcomes but also sets a new standard for equitable model performance in healthcare.

The flexibility and adaptability of FairDi make it applicable to a range of medical tasks, including classification and segmentation, underscoring its potential impact on the design of fair AI systems in critical domains. As healthcare increasingly adopts AI solutions, FairDi stands as a promising pathway toward ethical, inclusive, and high-performing models that prioritize equitable outcomes for all patients. We hope that this work paves the way for further advancements in fairness-centered AI, inspiring new strategies that embrace the complexities of fairness without sacrificing performance.

%\textcolor{red}{We need to increase the number of citations to at least 60.}
% \input{sec/6_finalcopy}
% \input{sec/7_remaining}
% \input{sec/8_formatting}
{
    \small
    \bibliographystyle{ieeenat_fullname}
    \bibliography{main}
}

\clearpage
\setcounter{page}{1}
\maketitlesupplementary

\section{Data Pre-processing}
\label{sec:data_preprocessing}
The data preprocessing in this work follows the approach outlined in~\cite{zong2022medfair}. For data splitting. The dataset is randomly partitioned into training, validation, and testing sets, maintaining an $80/10/10$ ratio unless stated otherwise. We then binarize both the prediction labels and sensitive attributes. Further details on this process are provided in the next section.

\section{Dataset Details}
\label{sec:dataset_details}

In this section, we provide details about the datasets. All datasets are publicly available and can be accessed through the URLs listed in Table~\ref{tab:dataset_access}. The dataset statistics are provided in Table~\ref{tab:dataset_statistics}. Additionally, we summarize the statistics for the subgroups and class labels in Tables~\ref{tab:ham10000_label_distribution} through \ref{tab:mimic_cxr_label_distribution}.

\subsection{Sensitive Attributes}

For the classification datasets, the sensitive attributes are defined as follows:
\begin{itemize}
    \item \textbf{Skin Type:} For the sensitive attribute of skin type in Fitzpatrick17k, we divided the data into two groups. The first group includes samples with skin types ranging from 0 to 2, while the second group consists of samples with skin types above 2.
    \item \textbf{Age:} For HAM10000, PAPILA, CheXpert, and MIMIC-CXR datasets, we classified the sensitive attribute of age into two distinct categories: the first group includes individuals aged 0 to 60, while the second group consists of those older than 60.
    \item \textbf{Race:} For CheXpert and MIMIC-CXR datasets, we defined two groups of race: a non-White group, comprising patients identified as 'Black' or 'Asian,' and a second group of patients identified as 'White'.
    \item \textbf{Gender:} For HAM10000, PAPILA, CheXpert, and MIMIC-CXR datasets, the sensitive attribute of gender has two categories: the first group consists of 'Male' patients, while the second group includes 'Female' patients.
\end{itemize}

The Harvard-FairSeg dataset~\cite{tian2024fairseg} encompasses six distinct sensitive attributes. Specifically, regarding racial demographics, it includes samples from three primary groups: Asian (919 samples), Black (1,473 samples), and White (7,608 samples). Gender distribution shows that females represent 58.5\% of subjects, with males comprising the remaining 41.5\%. Ethnically, 90.6\% of participants are Non-Hispanic, 3.7\% are Hispanic, and 5.7\% are unspecified. For preferred language, 92.4\% of individuals prefer English, 1.5\% prefer Spanish, 1\% prefer other languages, and 5.1\% have no specified language preference. Regarding marital status, 57.7\% are married or partnered, 27.1\% are single, 6.8\% are divorced, 0.8\% are legally separated, 5.2\% are widowed, and 2.4\% did not specify. In this study, we focus on the three sensitive attributes with the highest observed group-wise disparities: race, gender, and ethnicity.

\subsection{Disease  Labels}
\label{sec:disease_labels}

% \begin{itemize}
%     \item \textbf{Fitzpatrick17k:} The first group consists of samples labeled as malignant, while the second group includes samples with all other remaining labels.
%     \item \textbf{HAM10000:} We transformed the original labels into two categories: benign and malignant. The benign category includes samples with the original labels ``bcc'', ``bk1'', ``dermatofibroma'', ``nv'' and ``vasc''. The malignant category contains samples labeled as ``akiec'' and ``mel''.
%     \item \textbf{PAPILA:} We utilized samples from the ``healthy'' and ``glaucoma'' categories, excluding those from the ``suspect'' class.
%     \item \textbf{CheXpert \& MIMIC-CXR:} In the original dataset, each sample may correspond to one or more of the 14 labels. For our training, validation, and testing processes, we used the ``No-Finding'' label, as it is the only label suitable for binary classification.
% \end{itemize}

\begin{itemize}
    \item \textbf{Fitzpatrick17k:} We grouped the three partition labels into binary categories, namely \textit{benign} and \textit{malignant}. In this grouping, we classified both \textit{non-neoplastic} and \textit{benign} under the benign category, while \textit{malignant} remained as the malignant category. Furthermore, we used Fitzpatrick skin type labels as the sensitive attributes for our analysis.

    \item \textbf{HAM10000:} We followed the approach by~\cite{maron2019systematic} to categorize the seven diagnostic labels into two binary groups: \textit{benign} and \textit{malignant}. 
    \begin{itemize}
        \item The \textit{benign} group includes basal cell carcinoma (\texttt{bcc}), benign keratosis-like lesions (e.g., solar lentigines, seborrheic keratoses, and lichen-planus-like keratoses, \texttt{bkl}), dermatofibroma (\texttt{df}), melanocytic nevi (\texttt{nv}), and vascular lesions (angiomas, angiokeratomas, pyogenic granulomas, and hemorrhage, \texttt{vasc}).
        \item The \textit{malignant} group consists of actinic keratoses and intraepithelial carcinoma (Bowen's disease, \texttt{akiec}) as well as melanoma (\texttt{mel}).
    \end{itemize}
    Images without recorded sensitive attributes were excluded from the dataset, leaving a total of 9,948 images.

    \item \textbf{PAPILA:} In this dataset, we excluded the \textit{suspect} label class and focused on a binary classification task, using only images labeled as either \textit{glaucomatous} or \textit{non-glaucomatous}. The dataset contains both right-eye and left-eye images from the same patients. To divide the dataset into training, validation, and test sets, we followed a specific ratio: 70\% for training, 10\% for validation, and 20\% for testing. Importantly, we ensured that images from the same patient are not shared across these splits. This practice preserves the independence of the data subsets, which is essential for accurately evaluating the model’s performance.

    \item \textbf{CheXpert:} This dataset comprises 224,316 chest radiographs from 65,240 patients. Each image may be assigned one or more labels from a set of 14, representing various clinical observations. During preprocessing, we excluded images that lacked sensitive attribute labels. The \textit{No Finding} label is utilized for both training and testing purposes. All available frontal and lateral images were included, ensuring that images from the same patient were not shared across the training, validation, and test sets, following the approach outlined by~\cite{zong2022medfair}.

    \item \textbf{MIMIC-CXR:} Race data is sourced from the MIMIC-IV dataset~\cite{johnson2019mimic}, which is hosted in the PhysioNet database~\cite{goldberger2000physiobank}. We merged this data with the original MIMIC-CXR metadata using the \textit{subject ID} as the key. Other preprocessing steps closely follow the procedures used in the CheXpert dataset.
\end{itemize}

% Table 1: Dataset Access Information
\begin{table}[t!]
\centering
\caption{Access information for datasets used in the experiments.}
\scalebox{0.55}{
\begin{tabular}{@{}lp{0.7\textwidth}@{}}
\toprule
\textbf{Dataset} & \textbf{Access} \\ 
\midrule
Fitzpatrick17k & \url{https://github.com/mattgroh/fitzpatrick17k} \\ 
HAM10000 & \url{https://dataverse.harvard.edu/dataset.xhtml?persistentId=doi:10.7910/DVN/DBW86T} \\ 
PAPILA & \url{https://www.nature.com/articles/s41597-022-01388-1\#Sec6} \\ 
CheXpert & Original data: \url{https://stanfordmlgroup.github.io/competitions/chexpert/} \\ 
& Demographic data: \url{https://stanfordaimi.azurewebsites.net/datasets/192ada7c-4d43-466e-b8bb-b81992bb80cf} \\ 
MIMIC-CXR & \url{https://physionet.org/content/mimic-cxr-jpg/2.0.0/} \\ 
\bottomrule
\end{tabular}
}
\label{tab:dataset_access}
\end{table}

\vspace{1cm}

% Table 2: Dataset Statistics
\begin{table}[t!]
\centering
\caption{Dataset statistics including imaging modality, the number of images, and sensitive attributes.}
\scalebox{0.65}{
\begin{tabular}{@{}llll@{}}
\toprule
\textbf{Dataset} & \textbf{Imaging Modality} & \textbf{Number of Images} & \textbf{Sensitive Attribute} \\ 
\midrule
Fitzpatrick17k & Skin Dermatology (2D) & 16,012 & Skin Type \\ 
HAM10000 & Skin Dermatology (2D) & 9,948 & Age, Gender \\ 
PAPILA & Fundus Image (2D) & 420 & Age, Gender \\ 
CheXpert & Chest Radiographs (2D) & 222,793 & Age, Gender, Race \\ 
MIMIC-CXR & Chest Radiographs (2D) & 370,955 & Age, Gender, Race \\ 
\bottomrule
\end{tabular}
}
\label{tab:dataset_statistics}
\end{table}

% Table 5
\begin{table}[t!]
    \centering
    \caption{Label Distribution by Sensitive Attribute for the HAM10000 Dataset}
    \scalebox{0.85}{
    \begin{tabular}{l c c c}
        \toprule
        \textbf{Attribute} & \textbf{Label 0 (Healthy)} & \textbf{Label 1 (Unhealthy)} & \textbf{Total} \\
        \midrule
        Male        & 4492 & 908  & 5400 \\
        Female      & 4018 & 530  & 4548 \\
        \hline
        Age Group 0 & 6465 & 684  & 7149 \\
        Age Group 1 & 2045 & 754  & 2799 \\
        \bottomrule
    \end{tabular}
    }
    \label{tab:ham10000_label_distribution}
\end{table}

% Table 6
\begin{table}[t!]
    \centering
    \caption{Label Distribution by Sensitive Attribute for the PAPILA Dataset}
    \scalebox{0.85}{
    \begin{tabular}{l c c c}
        \toprule
        \textbf{Attribute} & \textbf{Label 0 (Healthy)} & \textbf{Label 1 (Unhealthy)} & \textbf{Total} \\
        \midrule
        Male        & 111 & 35  & 146 \\
        Female      & 222 & 52  & 274 \\
        \hline
        Age Group 0 & 163 & 15  & 178 \\
        Age Group 1 & 170 & 72  & 242 \\
        \bottomrule
    \end{tabular}
    }
    \label{tab:papila_label_distribution}
\end{table}

% Table 7
\begin{table}[t!]
    \centering
    \caption{Label Distribution by Sensitive Attribute for the Fitzpatrick17k Dataset}
    \scalebox{0.85}{
    \begin{tabular}{l c c c}
        \toprule
        \textbf{Attribute} & \textbf{Label 0 (Healthy)} & \textbf{Label 1 (Unhealthy)} & \textbf{Total} \\
        \midrule
        Skin Type 0 & 9412 & 1651 & 11063 \\
        Skin Type 1 & 4440 & 509  & 4949 \\
        \bottomrule
    \end{tabular}
    }
    \label{tab:fitzpatrick17k_label_distribution}
\end{table}

% Table 8
\begin{table}[t!]
    \centering
    \caption{Label Distribution by Sensitive Attribute for the CheXpert Dataset}
    \scalebox{0.75}{
    \begin{tabular}{l c c c}
        \toprule
        \textbf{Attribute} & \textbf{Label 0 (Healthy)} & \textbf{Label 1 (Unhealthy)} & \textbf{Total} \\
        \midrule
        Male        & 13,062 & 119,143 & 132,205 \\
        Female      & 9,177  & 81,411  & 90,588  \\
        \hline
        Age Group 0 & 16,274 & 90,488  & 106,762 \\
        Age Group 1 & 7,160  & 108,871 & 116,031 \\
        \hline
        White       & 12,538 & 113,095 & 125,633 \\
        Non-White   & 9,794  & 87,366  & 97,160  \\
        \bottomrule
    \end{tabular}
    }
    \label{tab:chexpert_label_distribution}
\end{table}

% Table 9
\begin{table}[t!]
    \centering
    \caption{Label Distribution by Sensitive Attribute for the MIMIC-CXR Dataset}
    \scalebox{0.75}{
    \begin{tabular}{l c c c}
        \toprule
        \textbf{Attribute} & \textbf{Label 0 (Healthy)} & \textbf{Label 1 (Unhealthy)} & \textbf{Total} \\
        \midrule
        Male        & 72,297  & 121,168 & 193,465 \\
        Female      & 76,113  & 101,377 & 177,490 \\
        \hline
        Age Group 0 & 63,072  & 45,407  & 108,479 \\
        Age Group 1 & 72,419  & 190,057 & 262,476 \\
        \hline
        White       & 78,193  & 146,403 & 224,596 \\
        Non-White   & 70,211  & 76,148  & 146,359 \\
        \bottomrule
    \end{tabular}
    }
    \label{tab:mimic_cxr_label_distribution}
\end{table}

\section{Fairness Measures}
\label{sec:fairness_measures}

For the classification metrics, the ES-AUC provides an equity-focused assessment by scaling the overall AUC based on performance differences across subgroups, calculated as follows:
\begin{equation}
\mathsf{ES-AUC} = \frac{\mathsf{AUC}(\mathcal{D})}{1 + \frac{1}{A} \sum_{g \in \mathcal{A}} \left| \mathsf{AUC}(\mathcal{D}) - \mathsf{AUC}(\mathcal{D}_g) \right|}
  \label{eq:ES-AUC}
\end{equation}
where $\mathsf{AUC}(\mathcal{D})$ represents the overall AUC for the whole test set, 
$g \in \mathcal{A} \subset \{0,\cdots,A\}$ represents 
the value of the identity attribute (e.g., for the gender attribute, the values can be male or female), $\mathsf{AUC}(\mathcal{D}_g)$ represents  the cohort-specific AUC on the test set $\mathcal{D}_g = \{(x,y,a) | (x,y,a) \in \mathcal{D}, a=g\}$ defined in~\eqref{eq:optimization_teacher}. This measure promotes fairness by proportionally reducing the overall AUC in response to increased subgroup disparities.

The Performance-Scaled Disparity (PSD) evaluates fairness by comparing cohort performance relative to overall AUC. Mean PSD calculates the standard deviation of AUC scores across cohorts, while Max PSD calculates the maximum absolute difference between cohorts' AUCs, both scaled by the overall AUC. These metrics are defined as follows:
\begin{equation}
\mathsf{Mean PSD}=\frac{\sqrt{\frac{1}{A}\sum_{g\in \mathcal{A}}\left(\mathsf{AUC}(\mathcal{D}_g)-\mathsf{AUC}_{\text{mean}}\right)^2}}{\mathsf{AUC}(\mathcal{D})},
  \label{eq:Mean-PSD}
\end{equation}
where $\mathsf{AUC}_{\text{mean}} = \frac{1}{A} \sum_{g\in \mathcal{A}} \mathsf{AUC}(\mathcal{D}_g)$, and 
\begin{equation}
\mathsf{Max PSD} =\frac{\left( \max_{g \in \mathcal{A}}(\mathsf{AUC}(\mathcal{D}_g))-\min_{g \in \mathcal{A}}(\mathsf{AUC}(\mathcal{D}_g))\right)}{\mathsf{AUC}(\mathcal{D})}
  \label{eq:Max-PSD}.
\end{equation}
These PSD measures provide interpretable insights into the model’s fairness performance, especially in clinical contexts where consistent accuracy across demographic groups is essential.

For the segmentation measures, the Dice coeffcient is calculated as~\cite{dice1945measures,sorensen1948method} 
\begin{equation}
\mathsf{Dice}(y,\hat{y}) = \frac{2 \times |y \cap \hat{y}|}{|y| + |\hat{y}|}, 
\end{equation}
where \(y\) and \(\hat{y}\) represent the actual and predicted segmented pixels, respectively. 
The Intersection over Union (IoU)~\cite{jaccard1901etude} is given by 
\begin{equation}
\mathsf{IoU}(y,\hat{y}) = \frac{|y \cap \hat{y}|}{|y \cup \hat{y}|}.
\end{equation}
Both metrics range from 0 to 1, with 1 indicating perfect overlap, but Dice tends to be slightly more sensitive to small overlaps than IoU due to its formula.
To adapt Dice to penalize discrepancies in performance across demographic groups, we first define a performance discrepancy as follows:
\begin{equation}
    \Delta(y,\hat{y}) = \sum_{g \in \mathcal{A}} \left | \mathsf{Dice}(y,\hat{y}) - \mathsf{Dice}(y,\hat{y};g)  \right |,
    \label{eq:Delta}
\end{equation}
where $\mathsf{Dice}(y,\hat{y};g)$ measures Dice for samples belonging to a particular value $g \in \mathcal{A}$.
The metric $\Delta(y,\hat{y})$ quantifies the deviation in performance of each demographic group relative to the overall performance, approaching zero when all groups achieve comparable segmentation accuracy.
To evaluate fairness across different groups, we compute the relative disparity between the overall segmentation accuracy and the accuracy within each demographic group. The Equity-Scaled Dice (ES-Dice) metric is then defined as:
\begin{equation}
    \mathsf{ES-Dice}(y,\hat{y}) = \frac{\mathsf{Dice}(y,\hat{y})}{1 + \Delta(y,\hat{y})}.
    \label{eq:ES-Dice}
\end{equation}
This formulation ensures that $\mathsf{ES-Dice}(.)$ is always less than or equal to 1. As $\Delta$ decreases (indicating more equitable segmentation performance across groups), $\mathsf{ES-Dice}(.)$ converges to $\mathsf{Dice}(.)$. Conversely, a higher $\Delta$ reflects greater disparity in segmentation performance across demographics, leading to a lower $\mathsf{ES-Dice}(.)$ score. This approach enables us to assess segmentation models not only for accuracy (measured by Dice) but also for fairness across demographic groups.
Note that the calculation of $\mathsf{ES-IoU}(.)$ follows the same steps as defined in~\eqref{eq:Delta} and~\eqref{eq:ES-Dice}.

\section{Complete Set of Results}
\label{sec:all_detailed_results}

\cref{tab:comparison_results_complete} shows the complete set of classification results, while~\cref{tab:fairseg_race,tab:fairseg_ethnicity,tab:fairseg_gender} show a complete set of segmentation results.

% Start of the table
\begin{table*}[t!]
    \centering
        \caption{Complete evaluation of fairness \textbf{classification} models across medical imaging benchmarks: HAM10000, Fitzpatrick17k, PAPILA, CheXpert, and MIMIC-CXR. We report overall AUC, subgroup-minimum AUC, ES-AUC, AUC gap, MeanPSD, and MaxPSD. Best results are highlighted.  
        %\textcolor{red}{and all measures in \%? I think based on the Mean PSD and Max PSD formulation, they are not in \%. The values for Mean PSD and Max PSD are not in percentage; they are calculated as proportions relative to the overall AUC. If we want to express them as percentages, we can simply multiply each value by 100. We should update this table based on adding new STOA methods.} 
    \label{tab:comparison_results_complete}}
    \scalebox{0.59}{
    \begin{tabular}{l c c c c c c c}

\toprule
\textbf{Dataset} & \textbf{Attr.} & \textbf{Metric} & \textbf{ERM} & \textbf{GroupDRO} & \textbf{SWAD} & \textbf{FIS} & \textbf{FairDi (Ours)} \\ 
\midrule
% \endhead

% \midrule \multicolumn{8}{r}{{Continued on next page}} \\ \midrule
% \endfoot

% \bottomrule
% \endlastfoot

% HAM10000 data
\multirow{12}{*}{HAM10000} & \multirow{6}{*}{Age} & Overall AUC↑ & 0.9000 & 0.8932 & 0.8961 & 0.9220 & \textbf{0.9447} \\ 
& & Min. AUC↑ & 0.7753 & 0.8000 & 0.8666 & 0.8832 & \textbf{0.9266} \\ 
& & ES-AUC↑ & 0.8383 & 0.8386 & 0.8712 & 0.9045 & \textbf{0.9353} \\ 
& & AUC Gap↓ & 0.1472 & 0.1301 & 0.0572 & 0.0388 & \textbf{0.0201} \\ 
& & Mean PSD↓ & $8.18 \times 10^{-2}$ & $7.28 \times 10^{-2}$ & $3.19 \times 10^{-2}$ & $2.42 \times 10^{-2}$ & $\mathbf{1.06 \times 10^{-2}}$ \\ 
& & Max PSD↓ & $1.64 \times 10^{-1}$ & $1.46 \times 10^{-1}$ & $6.38 \times 10^{-2}$ & $4.85 \times 10^{-2}$ & $\mathbf{2.13 \times 10^{-2}}$ \\ 
\cmidrule(lr){2-8}

& \multirow{6}{*}{Gender} & Overall AUC↑ & 0.8520 & 0.8703 & 0.9144 & 0.9300 & \textbf{0.9560} \\ 
& & Min. AUC↑ & 0.8312 & 0.8592 & 0.9114 & 0.9276 & \textbf{0.9553} \\ 
& & ES-AUC↑ & 0.8381 & 0.8625 & 0.9124 & 0.9283 & \textbf{0.9553} \\ 
& & AUC Gap↓ & 0.0331 & 0.0180 & 0.0044 & 0.0011 & \textbf{0.000029} \\ 
& & Mean PSD↓ & $1.94 \times 10^{-2}$ & $1.03 \times 10^{-2}$ & $2.41 \times 10^{-3}$ & $5.77 \times 10^{-4}$ & $\mathbf{1.52 \times 10^{-5}}$ \\ 
& & Max PSD↓ & $3.88 \times 10^{-2}$ & $2.07 \times 10^{-2}$ & $4.81 \times 10^{-3}$ & $1.15 \times 10^{-3}$ & $\mathbf{3.03 \times 10^{-5}}$ \\ 
\midrule

% Fitzpatrick17k data
\multirow{6}{*}{Fitzpatrick17k} & \multirow{6}{*}{Skin Type} & Overall AUC↑ & 0.9151 & 0.9198 & 0.9331 & 0.9023 & \textbf{0.9417} \\ 
& & Min. AUC↑ & 0.9067 & 0.8997 & 0.9094 & 0.8956 & \textbf{0.9406} \\ 
& & ES-AUC↑ & 0.8985 & 0.8943 & 0.9026 & 0.8895 & \textbf{0.9408} \\ 
& & AUC Gap↓ & 0.0370 & 0.0571 & 0.0675 & 0.0287 & \textbf{0.0020} \\ 
& & Mean PSD↓ & $2.02 \times 10^{-2}$ & $3.10 \times 10^{-2}$ & $3.62 \times 10^{-2}$ & $1.59 \times 10^{-2}$ & $\mathbf{1.10 \times 10^{-3}}$ \\ 
& & Max PSD↓ & $4.04 \times 10^{-2}$ & $6.21 \times 10^{-2}$ & $7.23 \times 10^{-2}$ & $3.17 \times 10^{-2}$ & $\mathbf{2.21 \times 10^{-3}}$ \\ 
\midrule

% PAPILA data
\multirow{12}{*}{PAPILA} & \multirow{6}{*}{Age} & Overall AUC↑ & 0.6498 & 0.7565 & 0.8227 & 0.9437 & \textbf{0.9807} \\ 
& & Min. AUC↑ & 0.5322 & 0.5826 & 0.6639 & 0.9168 & \textbf{0.9806} \\ 
& & ES-AUC↑ & 0.5779 & 0.6259 & 0.7043 & 0.9194 & \textbf{0.9713} \\ 
& & AUC Gap↓ & 0.2490 & 0.4174 & 0.3361 & 0.0528 & \textbf{0.0194} \\ 
& & Mean PSD↓ & $1.92 \times 10^{-1}$ & $2.76 \times 10^{-1}$ & $2.04 \times 10^{-1}$ & $2.80 \times 10^{-2}$ & $\mathbf{9.89 \times 10^{-3}}$ \\ 
& & Max PSD↓ & $3.83 \times 10^{-1}$ & $5.52 \times 10^{-1}$ & $4.09 \times 10^{-1}$ & $5.60 \times 10^{-2}$ & $\mathbf{1.98 \times 10^{-2}}$ \\ 
\cmidrule(lr){2-8}

& \multirow{6}{*}{Gender} & Overall AUC↑ & 0.7840 & 0.7969 & 0.7666 & 0.9391 & \textbf{0.9572} \\ 
& & Min. AUC↑ & 0.7804 & 0.7917 & 0.6667 & 0.9186 & \textbf{0.9508} \\ 
& & ES-AUC↑ & 0.7815 & 0.7921 & 0.7096 & 0.9259 & \textbf{0.9408} \\
& & AUC Gap↓ & \textbf{0.0008} & 0.0122 & 0.1608 & 0.0285 & 0.0349 \\ 
& & Mean PSD↓ & $\mathbf{5.10 \times 10^{-4}}$ & $7.65 \times 10^{-3}$ & $1.05 \times 10^{-1}$ & $1.52 \times 10^{-2}$ & $1.82 \times 10^{-2}$ \\ 
& & Max PSD↓ & $\mathbf{1.02 \times 10^{-3}}$ & $1.53 \times 10^{-2}$ & $2.10 \times 10^{-1}$ & $3.03 \times 10^{-2}$ & $3.64 \times 10^{-2}$ \\ 
\midrule

% CheXpert data
\multirow{18}{*}{CheXpert} & \multirow{6}{*}{Age} & Overall AUC↑ & 0.8809 & 0.8493 & \textbf{0.8875} & 0.8727 & 0.8735 \\ 
& & Min. AUC↑ & 0.7836 & 0.7991 & 0.8454 & 0.8464 & \textbf{0.8502} \\ 
& & ES-AUC↑ & 0.8376 & 0.8240 & \textbf{0.8667} & 0.8611 & 0.8614 \\ 
& & AUC Gap↓ & 0.1033 & 0.0613 & 0.0479 & \textbf{0.0256} & 0.0281 \\ 
& & Mean PSD↓ & $5.86 \times 10^{-2}$ & $3.61 \times 10^{-2}$ & $2.70 \times 10^{-2}$ & $\mathbf{1.47 \times 10^{-2}}$ & $1.61 \times 10^{-2}$ \\ 
& & Max PSD↓ & $1.17 \times 10^{-1}$ & $7.22 \times 10^{-2}$ & $5.40 \times 10^{-2}$ & $\mathbf{2.93 \times 10^{-2}}$ & $3.22 \times 10^{-2}$ \\ 
\cmidrule(lr){2-8}

& \multirow{6}{*}{Gender} & Overall AUC↑ & 0.8809 & 0.8627 & \textbf{0.8879} & 0.8752 & 0.8797 \\ 
& & Min. AUC↑ & 0.8807 & 0.8624 & \textbf{0.8872} & 0.8705 & 0.8768 \\ 
& & ES-AUC↑ & 0.8808 & 0.8624 & \textbf{0.8873} & 0.8699 & 0.8765 \\ 
& & AUC Gap↓ & \textbf{0.0002} & 0.0006 & 0.0013 & 0.0121 & 0.0074 \\ 
& & Mean PSD↓ & $\mathbf{1.14 \times 10^{-4}}$ & $3.48 \times 10^{-4}$ & $7.32 \times 10^{-4}$ & $6.95 \times 10^{-3}$ & $4.24 \times 10^{-3}$ \\ 
& & Max PSD↓ & $\mathbf{2.27 \times 10^{-4}}$ & $6.95 \times 10^{-4}$ & $1.46 \times 10^{-3}$ & $1.39 \times 10^{-2}$ & $8.48 \times 10^{-3}$ \\ 
\cmidrule(lr){2-8}

& \multirow{6}{*}{Race} & Overall AUC↑ & 0.8792 & 0.8630 & \textbf{0.8875} & 0.8665 & 0.8788 \\ 
& & Min. AUC↑ & 0.8784 & 0.8599 & \textbf{0.8865} & 0.8616 & 0.8783 \\ 
& & ES-AUC↑ & 0.8785 & 0.8602 & \textbf{0.8866} & 0.8605 & 0.8784 \\ 
& & AUC Gap↓ & 0.0016 & 0.0065 & 0.0021 & 0.014 & \textbf{0.0010} \\ 
& & Mean PSD↓ & $9.10 \times 10^{-4}$ & $3.77 \times 10^{-3}$ & $1.18 \times 10^{-3}$ & $8.08 \times 10^{-3}$ & $\mathbf{5.82 \times 10^{-4}}$ \\ 
& & Max PSD↓ & $1.82 \times 10^{-3}$ & $7.53 \times 10^{-3}$ & $2.37 \times 10^{-3}$ & $1.62 \times 10^{-2}$ & $\mathbf{1.16 \times 10^{-3}}$ \\ 
\midrule

% MIMIC-CXR data
\multirow{18}{*}{MIMIC-CXR} & \multirow{6}{*}{Age} & Overall AUC↑ & 0.8640 & 0.8349 & 0.8708 & 0.8753 & \textbf{0.8793} \\ 
& & Min. AUC↑ & 0.8106 & 0.7862 & 0.8215 & 0.8469 & \textbf{0.8555} \\ 
& & ES-AUC↑ & 0.8414 & 0.8130 & 0.8491 & 0.8562 & \textbf{0.8629} \\ 
& & AUC Gap↓ & 0.0532 & 0.0538 & 0.0510 & 0.0446 & \textbf{0.0381} \\ 
& & Mean PSD↓ & $3.08 \times 10^{-2}$ & $3.22 \times 10^{-2}$ & $2.93 \times 10^{-2}$ & $2.55 \times 10^{-2}$ & $\mathbf{2.17 \times 10^{-2}}$ \\ 
& & Max PSD↓ & $6.16 \times 10^{-2}$ & $6.44 \times 10^{-2}$ & $5.86 \times 10^{-2}$ & $5.10 \times 10^{-2}$ & $\mathbf{4.33 \times 10^{-2}}$ \\ 
\cmidrule(lr){2-8}

& \multirow{6}{*}{Gender} & Overall AUC↑ & 0.8645 & 0.8589 & 0.8705 & 0.8681 & \textbf{0.8776} \\ 
& & Min. AUC↑ & 0.8562 & 0.8496 & 0.8623 & 0.8580 & \textbf{0.8689} \\ 
& & ES-AUC↑ & 0.8584 & 0.8520 & 0.8645 & 0.8607 & \textbf{0.8710} \\ 
& & AUC Gap↓ & 0.0141 & 0.0161 & \textbf{0.0139} & 0.0173 & 0.0152 \\ 
& & Mean PSD↓ & $8.16 \times 10^{-3}$ & $9.37 \times 10^{-3}$ & $\mathbf{7.98 \times 10^{-3}}$ & $9.96 \times 10^{-3}$ & $8.66 \times 10^{-3}$ \\ 
& & Max PSD↓ & $1.63 \times 10^{-2}$ & $1.87 \times 10^{-2}$ & $\mathbf{1.60 \times 10^{-2}}$ & $1.99 \times 10^{-2}$ & $1.73 \times 10^{-2}$ \\ 
\cmidrule(lr){2-8}

& \multirow{6}{*}{Race} & Overall AUC↑ & 0.8626 & 0.8565 & 0.8710 & 0.8787 & \textbf{0.8811} \\ 
& & Min. AUC↑ & 0.8552 & 0.8490 & 0.8638 & 0.8709 & \textbf{0.8714} \\ 
& & ES-AUC↑ & 0.8589 & 0.8531 & 0.8672 & \textbf{0.8742} & 0.8741 \\ 
& & AUC Gap↓ & 0.0085 & \textbf{0.0080} & 0.0088 & 0.0103 & 0.0160 \\ 
& & Mean PSD↓ & $4.93 \times 10^{-3}$ & $\mathbf{4.67 \times 10^{-3}}$ & $5.05 \times 10^{-3}$ & $5.86 \times 10^{-3}$ & $9.08 \times 10^{-3}$ \\ 
& & Max PSD↓ & $9.85 \times 10^{-3}$ & $\mathbf{9.34 \times 10^{-3}}$ & $1.01 \times 10^{-2}$ & $1.17 \times 10^{-2}$ & $1.82 \times 10^{-2}$ \\ 
% \midrule

% % Summary row
% \multicolumn{3}{c}{Avg. Overall AUC Score} & 84.85 & 85.11 & 87.35 & 89.76 & \textbf{91.37} \\ 
% \multicolumn{3}{c}{Avg. Min. AUC Score} & 80.82 & 81.27 & 83.50 & 88.15 & \textbf{90.50} \\ 
% \multicolumn{3}{c}{Avg. ES-AUC Score} & 82.64 & 82.53 & 84.74 & 88.64 & \textbf{90.62} \\ 
% \multicolumn{3}{c}{Avg. AUC Gap Score} & 5.89 & 7.10 & 6.83 & 2.49 & \textbf{1.66} \\ 
% \multicolumn{3}{c}{Avg. Mean PSD Score} & 0.0379 & 0.0440 & 0.0410 & 0.0141 & \textbf{0.0091} \\ 
% \multicolumn{3}{c}{Avg. Max PSD Score} & 0.0758 & 0.0881 & 0.0820 & 0.0282 & \textbf{0.0182} \\ 
% \midrule

% % Ranking row
% \multicolumn{3}{c}{Avg. Overall AUC Rank} & 3.55 & 4.45 & 2.55 & 2.91 & \textbf{1.55} \\ 
% \multicolumn{3}{c}{Avg. Min. AUC Rank} & 3.91 & 4.36 & 2.64 & 2.73 & \textbf{1.36} \\ 
% \multicolumn{3}{c}{Avg. ES-AUC Rank} & 3.82 & 4.45 & 2.45 & 2.73 & \textbf{1.55} \\ 
% \multicolumn{3}{c}{Avg. AUC Gap Rank} & 3.00 & 3.55 & 3.27 & 3.00 & \textbf{2.18} \\ 
% \multicolumn{3}{c}{Avg. Mean PSD Rank} & 3.00 & 3.55 & 3.27 & 3.00 & \textbf{2.18} \\ 
% \multicolumn{3}{c}{Avg. Max PSD Rank} & 3.00 & 3.55 & 3.27 & 3.00 & \textbf{2.18} \\ 
\bottomrule

\end{tabular}
}
\end{table*}

% % 
% Having the supplementary compiled together with the main paper means that:
% % 
% \begin{itemize}
% \item The supplementary can back-reference sections of the main paper, for example, we can refer to \cref{sec:intro};
% \item The main paper can forward reference sub-sections within the supplementary explicitly (e.g. referring to a particular experiment); 
% \item When submitted to arXiv, the supplementary will already included at the end of the paper.
% \end{itemize}
% % 
% To split the supplementary pages from the main paper, you can use \href{https://support.apple.com/en-ca/guide/preview/prvw11793/mac#:~:text=Delete%20a%20page%20from%20a,or%20choose%20Edit%20%3E%20Delete).}{Preview (on macOS)}, \href{https://www.adobe.com/acrobat/how-to/delete-pages-from-pdf.html#:~:text=Choose%20%E2%80%9CTools%E2%80%9D%20%3E%20%E2%80%9COrganize,or%20pages%20from%20the%20file.}{Adobe Acrobat} (on all OSs), as well as \href{https://superuser.com/questions/517986/is-it-possible-to-delete-some-pages-of-a-pdf-document}{command line tools}.

\begin{table*}[t!]
    \centering
    \caption{\textbf{Segmentation} performance for optic cup and rim on the Harvard-FairSeg dataset, with \textbf{race} as the sensitive attribute.}
    \scalebox{0.6}{
    \begin{tabular}{l l c c c c c c c c c c}
    \toprule
    & \textbf{Method} & \textbf{Overall ES-Dice↑} & \textbf{Overall Dice↑} & \textbf{Overall ES-IoU↑} & \textbf{Overall IoU↑} & \textbf{Asian Dice↑} & \textbf{Black Dice↑} & \textbf{White Dice↑} & \textbf{Asian IoU↑} & \textbf{Black IoU↑} & \textbf{White IoU↑} \\
    \midrule
    \multirow{5}{*}{\textbf{Cup}} 
    & TransUNet & 0.8281 & 0.8481 & 0.7300 & 0.7532 & 0.8270 & 0.8489 & 0.8503 & 0.7277 & \textbf{0.7576} & 0.7551 \\
    & TransUNet+ADV & 0.8256 & 0.8410 & 0.7265 & 0.7432 & 0.8246 & 0.8417 & 0.8426 & 0.7260 & 0.7482 & 0.7440 \\
    & TransUNet+GroupDRO & 0.8201 & 0.8442 & 0.7252 & 0.7479 & 0.8197 & 0.8469 & 0.8464 & 0.7232 & 0.7529 & 0.7495 \\
    & TransUNet+FEBS & 0.8253 & 0.8464 & 0.7265 & 0.7497 & 0.8248 & 0.8484 & 0.8484 & 0.7247 & 0.7550 & 0.7513 \\
    & \textbf{TransUNet+FairDi} & \textbf{0.8386} & \textbf{0.8532} & \textbf{0.7365} & \textbf{0.7543} & \textbf{0.8386} & \textbf{0.8527} & \textbf{0.8555} & \textbf{0.7359} & 0.7552 & \textbf{0.7591} \\
    \midrule
    \multirow{5}{*}{\textbf{Rim}} 
    & TransUNet & 0.7034 & 0.7927 & 0.5848 & 0.6706 & 0.7457 & 0.7307 & 0.8106 & 0.6160 & 0.5991 & 0.6913 \\
    & TransUNet+ADV & 0.7000 & 0.7906 & 0.5825 & 0.6682 & 0.7413 & 0.7286 & 0.8087 & 0.6116 & 0.5982 & 0.6888 \\
    & TransUNet+GroupDRO & 0.7002 & 0.7896 & 0.5814 & 0.6674 & 0.7470 & 0.7229 & 0.8080 & 0.6183 & 0.5899 & 0.6887 \\
    & TransUNet+FEBS & 0.7050 & 0.7950 & 0.5871 & 0.6725 & 0.7479 & 0.7325 & \textbf{0.8130} & 0.6185 & 0.6020 & \textbf{0.6935} \\
    & \textbf{TransUNet+FairDi} & \textbf{0.7278} & \textbf{0.8027} & \textbf{0.5994} & \textbf{0.6828} & \textbf{0.7533} & \textbf{0.7492} & 0.8027 & \textbf{0.6235} & \textbf{0.6128} & 0.6927 \\
    \bottomrule
    \end{tabular}
    }    
    \label{tab:fairseg_race}
\end{table*}

\begin{table*}[t!]
    \centering
    \caption{\textbf{Segmentation} performance for optic cup and rim on the Harvard-FairSeg dataset, with \textbf{gender} as the sensitive attribute.}
    \scalebox{0.6}{
    \begin{tabular}{l l c c c c c c c c}
    \toprule
    & \textbf{Method} & \textbf{Overall ES-Dice↑} & \textbf{Overall Dice↑} & \textbf{Overall ES-IoU↑} & \textbf{Overall IoU↑} & \textbf{Male Dice↑} & \textbf{Female Dice↑} & \textbf{Male IoU↑} & \textbf{Female IoU↑} \\
    \midrule
    \multirow{5}{*}{\textbf{Cup}} 
    & TransUNet & 0.8435 & 0.8481 & 0.7490 & 0.7532 & 0.8458 & 0.8513 & 0.7508 & 0.7564 \\
    & TransUNet+ADV & 0.8342 & 0.8345 & 0.7348 & 0.7356 & 0.8344 & 0.8348 & 0.7361 & 0.7350 \\
    & TransUNet+GroupDRO & 0.8405 & 0.8478 & 0.7453 & 0.7522 & 0.8441 & 0.8528 & 0.7483 & 0.7575 \\
    & TransUNet+FEBS & 0.8464 & 0.8489 & 0.7492 & 0.7530 & 0.8494 & 0.8514 & 0.7505 & 0.7556 \\
    & \textbf{TransUNet+FairDi} & \textbf{0.8521} & \textbf{0.8535} & \textbf{0.7576} & \textbf{0.7623} & \textbf{0.8529} & \textbf{0.8545} & \textbf{0.7577} & \textbf{0.7639} \\
    \midrule
    \multirow{5}{*}{\textbf{Rim}} 
    & TransUNet & 0.7882 & 0.7927 & 0.6659 & 0.6706 & 0.7951 & 0.7894 & 0.6736 & 0.6665 \\
    & TransUNet+ADV & 0.7754 & 0.7852 & 0.6522 & 0.6630 & 0.7905 & 0.7779 & 0.6699 & 0.6534 \\
    & TransUNet+GroupDRO & 0.7893 & 0.7917 & 0.6673 & 0.6699 & 0.7930 & 0.7900 & 0.6716 & 0.6677 \\
    & TransUNet+FEBS & 0.7851 & 0.7898 & 0.6655 & 0.6698 & 0.7924 & \textbf{0.7932} & 0.6678 & 0.6653 \\
    & \textbf{TransUNet+FairDi} & \textbf{0.7917} & \textbf{0.7988} & \textbf{0.6742} & \textbf{0.6783} & \textbf{0.8016} & 0.7926 & \textbf{0.6792} & \textbf{0.6731} \\
    \bottomrule
    \end{tabular}
    }    
    \label{tab:fairseg_gender}
\end{table*}

\begin{table*}[t!]
    \centering
    \caption{\textbf{Segmentation} performance for optic cup and rim on the Harvard-FairSeg dataset, with \textbf{ethnicity} as the sensitive attribute.}
    \scalebox{0.6}{
    \begin{tabular}{l l c c c c c c c c}
    \toprule
    & \textbf{Method} & \textbf{Overall ES-Dice↑} & \textbf{Overall Dice↑} & \textbf{Overall ES-IoU↑} & \textbf{Overall IoU↑} & \textbf{Hispanic Dice↑} & \textbf{Non-Hispanic Dice↑} & \textbf{Hispanic IoU↑} & \textbf{Non-Hispanic IoU↑} \\
    \midrule
    \multirow{5}{*}{\textbf{Cup}} 
    & TransUNet & 0.8281 & 0.8481 & 0.7300 & 0.7532 & 0.8463 & \textbf{0.8704} & 0.7508 & \textbf{0.7826} \\
    & TransUNet+ADV & 0.8112 & 0.8320 & 0.7083 & 0.7315 & 0.8304 & 0.8561 & 0.7294 & 0.7622 \\
    & TransUNet+GroupDRO & 0.8332 & 0.8482 & 0.7358 & 0.7526 & 0.8468 & 0.8648 & 0.7507 & 0.7735 \\
    & TransUNet+FEBS & 0.8320 & 0.8483 & 0.7359 & 0.7542 & 0.8501 & 0.8661 & 0.7515 & 0.7764 \\
    & \textbf{TransUNet+FairDi} & \textbf{0.8396} & \textbf{0.8516} & \textbf{0.7581} & \textbf{0.7683} & \textbf{0.8564} & 0.8611 & \textbf{0.7698} & 0.7803 \\
    \midrule
    \multirow{5}{*}{\textbf{Rim}} 
    & TransUNet & 0.7815 & 0.7927 & 0.6626 & 0.6706 & 0.7914 & \textbf{0.8057} & 0.6695 & 0.6815 \\
    & TransUNet+ADV & 0.7774 & 0.7841 & 0.6557 & 0.6602 & 0.7829 & 0.7915 & 0.6590 & 0.6658 \\
    & TransUNet+GroupDRO & \textbf{0.7904} & 0.7943 & 0.6672 & 0.6733 & 0.7936 & 0.7901 & 0.6728 & 0.6646 \\
    & TransUNet+FEBS & 0.7857 & 0.7939 & 0.6692 & 0.6754 & 0.7943 & 0.8040 & 0.6697 & 0.6789 \\
    & \textbf{TransUNet+FairDi} & 0.7871 & \textbf{0.7962} & \textbf{0.6755} & \textbf{0.6792} & \textbf{0.8068} & 0.7952 & \textbf{0.6784} & \textbf{0.6839} \\
    \bottomrule
    \end{tabular}
    }    
    \label{tab:fairseg_ethnicity}
\end{table*}

\end{document}